\documentclass[a4paper,11pt,oneside]{report}
\usepackage[MScThesis]{EPFLreport}
\usepackage{xspace}

\usepackage{amsmath}
\usepackage{wrapfig}
\usepackage{hyperref}

\usepackage{subcaption}

\usepackage{natbib}
\setcitestyle{authoryear,open={(},close={)}} 

\title{Implementing and Experimenting with Diffusion Models for Text-to-Image Generation}
\author{Robin Zbinden}
\supervisor{Dr. Luis Barba}
\adviser{Prof. Martin Jaggi}

\begin{document}
\maketitle
\makededication
\makeacks

\begin{abstract}

Taking advantage of the many recent advances in deep learning, text-to-image generative models currently have the merit of attracting the general public attention. Two of these models, DALL-E 2 and Imagen, have demonstrated that highly photorealistic images could be generated from a simple textual description of an image. Based on a novel approach for image generation called diffusion models, text-to-image models enable the production of many different types of high resolution images, where human imagination is the only limit. 

However, these models require exceptionally large amounts of computational resources to train, as well as handling huge datasets collected from the internet. In addition, neither the codebase nor the models have been released.
It consequently prevents the AI community from experimenting with these cutting-edge  models, making the reproduction of their results complicated, if not impossible.

In this thesis, we aim to contribute by firstly reviewing the different approaches and techniques used by these models, and then by proposing our own implementation of a text-to-image model. Highly based on DALL-E 2, we introduce several slight modifications to tackle the high computational cost induced. We thus have the opportunity to experiment in order to understand what these models are capable of, especially in a low resource regime. In particular, we provide additional and analyses deeper than the ones performed by the authors of DALL-E 2, including ablation studies.

Besides, diffusion models use so-called guidance methods to help the generating process. We introduce a new guidance method which can be used in conjunction with other guidance methods to improve image quality.
Finally, the images generated by our model are of reasonably good quality, without having to sustain the significant training costs of state-of-the-art text-to-image models.
\end{abstract}

\begin{frenchabstract}

Tirant parti des nombreuses avancées récentes en matière d'apprentissage profond, les modèles génératifs texte-à-image ont actuellement le mérite d'attirer l'attention du grand public. Deux de ces modèles, DALL-E 2 et Imagen, ont démontré que des images hautement photoréalistes pouvaient être générées à partir de la simple description textuelle d'une image. Basés sur une nouvelle approche de génération d'images appelée modèles de diffusion, les modèles texte-à-image permettent de produire de nombreux différents types d'images à haute résolution, où l'imagination humaine est la seule limite.

Toutefois, l'apprentissage de ces modèles nécessite des ressources computationnelles exceptionnellement importantes, ainsi que la manipulation d'énormes ensembles de données collectées sur internet. En outre, ni le code, ni les modèles n'ont été publiés. Cela empêche donc la communauté d'intelligence artificielle d'expérimenter ces modèles de pointe, rendant la reproduction de leurs résultats compliquée, voire impossible.

Dans cette thèse, nous souhaitons apporter notre contribution en passant d'abord en revue les différentes approches et techniques utilisées par ces modèles, puis en proposant notre propre implémentation d'un modèle texte-à-image. Fortement inspiré de DALL-E 2, nous introduisons plusieurs légères modifications pour faire face au coût de calcul élevé induit. Nous avons ainsi l'opportunité d'expérimenter afin de comprendre ce dont ces modèles sont capables, en particulier dans une position de faibles ressources. De plus, nous fournissons des analyses supplémentaires et plus profondes que celles effectuées par les auteurs de DALL-E 2, y compris des études ablatives.

En outre, les modèles de diffusion utilisent des méthodes dites de guidage pour aider le processus de génération. Nous introduisons une nouvelle méthode de guidage qui peut être utilisée en conjonction avec d'autres méthodes de guidage pour améliorer la qualité des images. Enfin, les images générées par notre modèle sont d'assez bonne qualité, sans avoir à supporter les coûts d’apprentissage importants des modèles texte-à-image de pointe.

\end{frenchabstract}

\maketoc

\chapter{Introduction}

Deep Learning models have recently shown their capacity and versatility to be applied to various, unstructured, and high-dimensional sorts of data. 
In contrast to traditional machine learning models, deep artificial neural networks can easily deal with tremendous amounts of data, being able to model the complex reality of the world.
Additionally, many concepts of deep learning are easily transferable among different modalities. All of this gives the possibility to use them for many applications including natural language translation \citep{languagetranslation}, detection of genetic disorders \citep{geneticdisorder}, and control of nuclear fusion plasma \citep{nuclearfusion}.
Moreover, an increasing number of researches are now focusing on the objective to find a unique neural architecture and systematic training procedure which could be applied to any source of data \citep{data2vec}. In this fruitful context, a new category of models called multimodal machine learning has emerged, aiming to jointly process multiple sources of data such as text and image.

We focus in this thesis on one type of multimodal models:  \textit{text-to-image generative models}. These models learn to synthesize images given an image description and recent works have shown that scaling up their size makes them able to produce complex photorealistic images \citep{dalle2, imagen}. They also have a \textit{zero-shot} learning capacity to generalize, which enables them to synthesize image types that have not been seen during training \citep{dalle}. However, training these models necessitates handling massive datasets of captioned images. For example, state-of-the-art models for text-to-image generation DALL-E 2 \citep{dalle2} and Imagen \citep{imagen} use 650M and 860M of image-text pairs respectively. It involves particularly long trainings, requiring important amounts of computational power and resources. In addition, the datasets and code repositories are often unavailable to the public community, which makes the process of replicating these generative models even more difficult.

\begin{figure}
    \centering
    \includegraphics[width=1\linewidth]{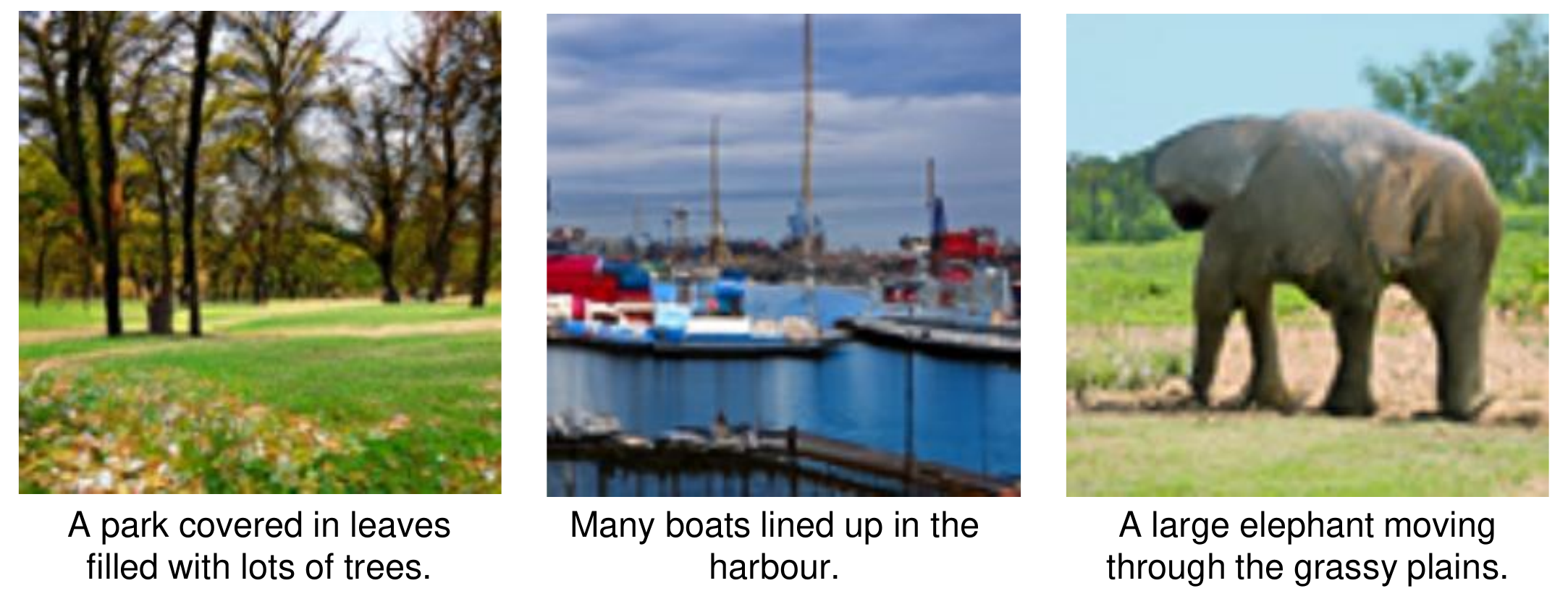}
    \caption{Some of the images synthesized by our text-to-image generative model, conditioned on the corresponding textual caption.}
    \label{fig:best_images}
\end{figure}

Moreover, while the pace of development is high, little work has been devoted to determine what these models are really capable of, partly due to the difficulty of reproduction. In addition, novel approaches use heuristics which often lack theoretical foundations. It is therefore crucial to allow the wide deep learning researchers community to experiment with these models, by finding techniques to replicate them more easily and more efficiently. 
This aspiration is then in line with the DALL-E mini model \citep{dalle_mini}, which intends to reproduce the text-to-image generative model DALL-E \citep{dalle}. In particular, it proposes several tricks to decrease the computational load, including a smaller dataset, the use of pre-trained models, and the replacement of the auto-regressive model by a bidirectional encoder. However, the approach followed by DALL-E and its mini counterpart has been outperformed by methods leveraging diffusion models to produce higher-quality images. Diffusion models are a new class of generative models which are starting to quietly replace previous models such as Generative Adversarial Networks (GANs) and Variational Autoencoders (VAEs). In particular, the two state-of-the-art methods in text-to-image generation DALL-E 2 and Imagen use them to synthesize high-resolution photorealistic images.

In this thesis, we propose therefore an implementation\footnote{Our implementation is available at \url{https://github.com/epfml/text_to_image_generation}} of a text-to-image generative model, based on DALL-E 2. Likewise DALL-E mini, we introduce slight modifications to make DALL-E 2 trainable without an explosion of computational load. This is, to our knowledge, the first available replication of a text-to-image generative model based on diffusion models. We then have the opportunity to experiment with this model to understand the possibilities and limitations of our approach. Its characteristics allow us to manipulate image and text representations in the form of vector embeddings capturing the semantic content of the data. It enables us to perform simple algebra operations on these vectors to gain a more detailed understanding of what information they contain. Finally, we propose a simple new guidance method called \textit{image guidance} to help the generating process by means of an extra image, and demonstrate its usefulness with some experiments.

This master's thesis is structured as follows. We start in \autoref{sec:related_work} by introducing the related work from which we take inspiration to build our method, as well as alternative approaches for text-to-image generation. Next, we provide in \autoref{sec:background} some technical background knowledge which is essential to understand our system. In particular, we take a deep dive into diffusion models and how to guide them to improve sample quality. We also take the time to explain in detail the CLIP model \citep{clip}, as our method uses it extensively. Chapter \ref{sec:method} simply presents our method and its different components, whereas \autoref{sec:experiments} proposes some experiments to assess the quality of our system and to provide insights of why this method is useful. We then discuss in \autoref{sec:discussion} the results obtained, by comparing them with the related works and proposing future directions for this work. We also include a section on the societal impacts caused by the type of models we use in our method. Finally, we conclude this thesis in \autoref{sec:conclusion} by reviewing its principal aspects.

\chapter{Related work}
\label{sec:related_work}

We describe in this chapter the different related works and approaches for text-to-image generation. We start in \autoref{sec:gen_models} by reviewing what deep generative models are and we briefly explain the different approaches devised to generate samples from a distribution. We then explore text-to-image generation and in the process we do a succinct recall of computer vision and natural language processing models.

\section{Deep Generative Models}
\label{sec:gen_models}

Machine learning models are divided into two distinct categories: discriminative and generative models. Discriminative models aim to approximate $p(y|x)$, allowing to predict a target $y$ given the observation $x$. Logistic regression and decision tree models fall into this category, but also some deep neural networks, for example image classifier models. In contrast, generative modelling aims at solving a more general task consisting of learning the joint distribution $p(x)$ over all the variables $x$. $p(x)$ can be for instance the distribution of the pixels on an image. This is a more complex task, but it allows drawing samples from the distribution $p(x)$, e.g., to generate images. A good generative model $p_\theta(x)$ therefore tries to maximize the likelihood of the data, or at least an approximation of the likelihood when its computation is not tractable.
Moreover, these models can be easily conditioned on some value $y$ to obtain a conditional probability distribution $p(x|y)$. As an example, $y$ can be a label or a caption indicating the content of the image $x$. 

Over the past years, different approaches and models relying on deep learning have been devised to generate samples, including GANs, VAEs, flow-based models, autoregressive models and finally diffusion models.
Since we will often mention them and their performances, we provide here a brief description for each of them, as well as their upsides and downsides:
\begin{itemize}
    \item \textbf{Generative Adversarial Networks (GANs)}: GANs \citep{GAN} are composed in general of two distinct models which have antagonistic objectives. The first is called the generative model $G$ and tries to capture the data distribution, while the second named the discriminative model $D$ is designed to differentiate synthetic samples generated by $G$ from real samples of the data distribution. In other words, $G$ tries to fool $D$ during this adversarial training. A lot of work focusing on models from the GANs family were produced in the past years. Even if the sample quality is generally great, they suffer from mode collapse leading to a lack of sample diversity \citep{biasGAN}. Moreover, the adversarial nature of the training makes this latter unstable and therefore often laborious \citep{unstableGAN}.
    \item \textbf{Variational Auto-Encoders (VAEs):} VAEs \citep{VAE} are probabilistic generative models learning first to encode data into a constrained lower-dimensional latent space and second to decode from the latter to get the data back. The constraints on the latent space enable to obtain samples close to the original data distribution, by simply randomly sampling from the latent space.
    VAEs often obtain high log-likelihood values, but struggle to produce non-blurry high quality samples.
    \item \textbf{Flow-based generative models:} This family of models consists of applying a sequence of invertible parameterized functions \citep{nice, normflow}. In contrast to GANs and VAEs, flow-based generative models maximize directly the exact log-likelihood probability. Nonetheless, these models are generally outperformed in terms of sample quality for now, primarily because of the difficulty of finding effective invertible architectures.
    \item \textbf{Autoregressive models:} Autoregressive models generate the output sequentially, conditioned on past parts of the output. For instance for image generation, PixelCNN \citep{pixelCNN} and iGPT \citep{iGPT} generate the image pixel by pixel. However, their autoregressive nature makes the sampling complexity grow linearly with the size of the output.
    \item \textbf{Diffusion models:} Diffusion models demonstrated recently their ability to produce high quality samples. They generate samples from a data distribution by progressively removing noise from a noisy data sample. This is done by sequentially applying the same model to the data. In consequence, the principal downside is the large sampling time, consisting of many Deep Neural Network (DNN) forward passes. Nevertheless, less expensive alternatives have been devised to reduce this number of forward passes \citep{DDIM}. We explain in more details the mechanisms of diffusion models in the next section.
\end{itemize}
Note that these approaches are not exclusive, and some recent models combine concepts such as VQ-GANs \citep{vqgan}, autoregressive diffusion models \citep{ARDM}, or denoising diffusion GANs \citep{DDGAN}. 

\section{Text-to-Image Generative Models}

Now that we have a clear view about what generative models are and that we know a few instances of them, we delve into the literature of a specific type of generative models which are text-to-image models. As their name suggests, they aim to synthesize images from text, involving then two different modalities. It entails to process the data differently by using a particular sort of model depending on the type of data. Image data requires models belonging to the field of computer vision (CV), whereas text data is linked to the field of natural language processing (NLP). We therefore start by briefly summarizing the recent advances in these two areas, before introducing state-of-the-art models in text-to-image generation in a second phase.

Since the seminal work of Yann LeCun \citep{LeCun-gradient}, Deep Neural Architectures (DNNs) have revolutionised the area of \textbf{computer vision}, being able to obtain prodigious levels of performance on a diverse set of visual tasks, including image classification \citep{AlexNet}, image generation \citep{glow}, and object detection \citep{YOLO}.
While Convolutional Neural Networks (CNNs) have long been in the past years the \textit{de facto} favourite backbone architectures in computer vision, new types of models have recently emerged. An architecture imported and adapted from NLP called the Visual Transformer \citep{vit} achieved similar and even sometimes better results than traditional CNN models on different CV applications. Despite this paradigm shift, CNNs have not yet been abandoned \citep{convnext}, and hybrid architectures are being developed to combine the built-in inductive biases of the convolution operation with the self-attention mechanisms of Transformer models \citep{conv_transformer}. Moreover, last improvements in computer vision are principally due to the use of larger models and bigger datasets. They were made possible by the rise of self-supervised learning (again coming from NLP), allowing to leverage massive unlabeled datasets from the web \citep{vit}.

On the other hand, the boom of efficient and valuable models for \textbf{natural language processing} took place a few years after the one in computer vision. The expansion of language models arose with the development of the Transformer architecture mentioned above \citep{transformer}, replacing recurrent neural networks (RNNs). Foundation models implementing the Transformer such as BERT \citep{bert} or GPT \citep{gpt} have demonstrated that profound natural language understanding could emerge using generative pre-training along with self-supervised learning. It consists of first pre-training the language model on a various corpus of unlabeled text, and secondly fine-tuning it on a more specific downstream task, without substantial architecture alterations. It hence enables to transfer the richness of vast datasets to a broad range of tasks, by learning a vector representation of a piece of text. These vector representations, also called text embeddings, are supposed to capture the syntactic and semantic patterns in natural language. They are then very effective for many language tasks. In addition and similarly to computer vision, scaling the size of the models and datasets led to major improvements. The third and biggest version of GPT \citep{GPT3} showed that a larger generative language model could handle more difficult reasoning tasks. Likewise, the largest language model ever created, PaLM (540B parameters), demonstrated its remarkable adaptability, being even able to explain jokes to humans \citep{palm}. 

\begin{figure}
    \centering
    \includegraphics[width=1\linewidth]{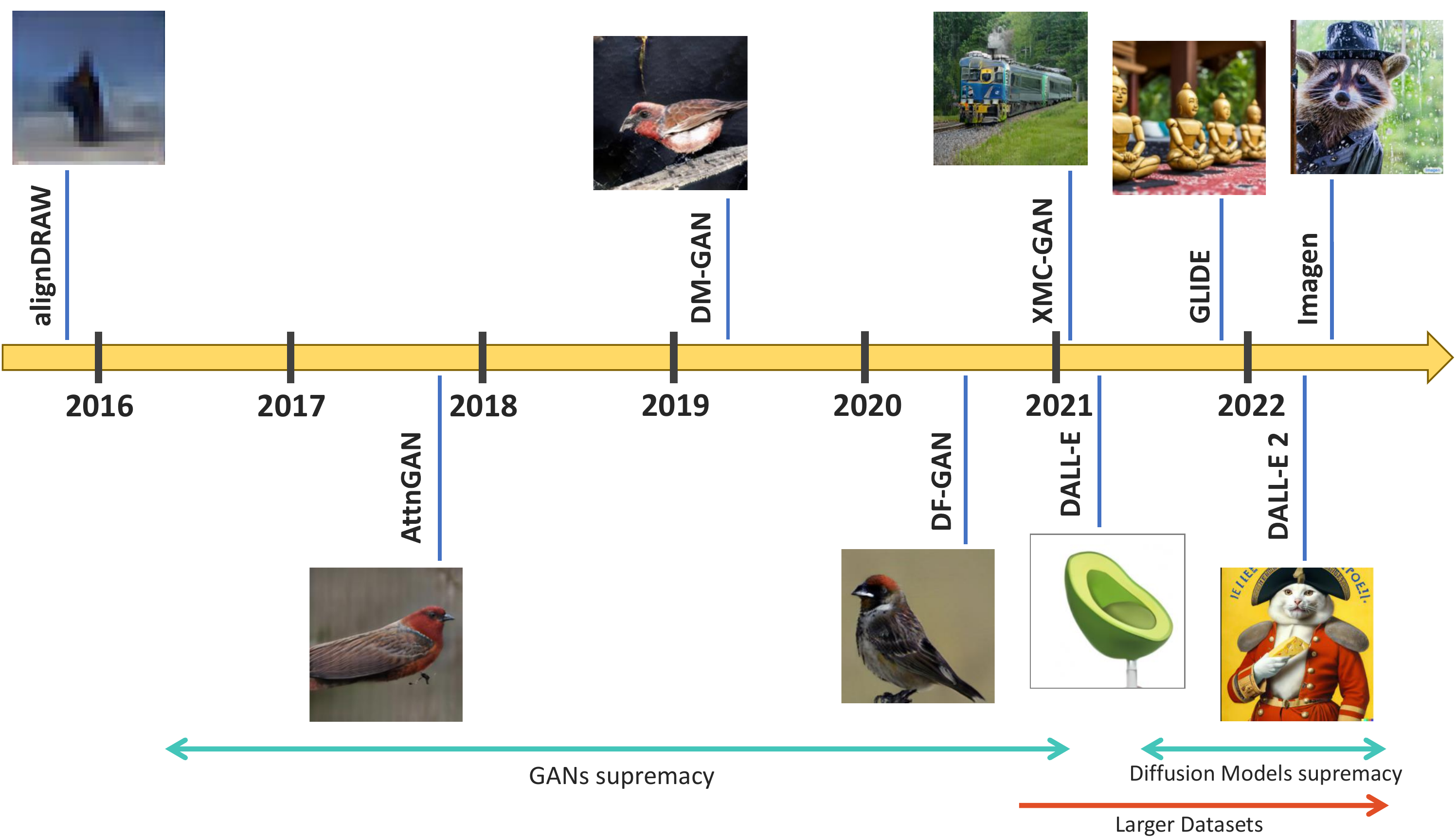}
    \caption{Timeline of text-to-image generative models.}
    \label{fig:timeline}
\end{figure}

Therefore, the astounding effectiveness of deep learning in computer vision and natural language processing naturally gave birth to multimodal models combining these two fields, i.e., \textbf{vision-language models}. They are able to learn joint representations of text and image to accomplish many vision-language tasks including Visual Question Answering (VQA), Visual Retrieval (VR), and Visual Captioning (VC). In particular, we focus on the text-to-image generation task as this is the one we are interested in. The idea of synthesizing an image given a caption began with the early work alignDRAW \citep{alignDRAW}. However, the image quality was very low and the scenes and objects generated were barely recognizable. This was followed by five years of modest improvements instigated by the progresses of GANs, with models such as AttnGAN \citep{AttnGAN}, DM-GAN \citep{DM-GAN}, or DF-GAN \citep{DF-GAN}. Although some of the content of the captions was beginning to be represented, the images were still not realistic, except for restricted and simple datasets, e.g., the CUB dataset \citep{cub} which only consists of bird images. Integrating contrastive learning in the pipeline and especially increasing the dataset size enabled XMC-GAN \citep{XMC-GAN} to produce better images depicting clearer scenes. 
In the same vein, authors of DALL-E \citep{dalle} showed that further scaling up the dataset to 250M image-text pairs could enable zero-shot learning. 
Thus, the DALL-E model is able to mix different objects, concepts, and places to produce non-iconic images, e.g., an avocado chair. 
DALL-E however doesn't use GANs and instead exploits a VQ-VAE \citep{vqvae}, as well as two Transformers, riding on the wave of the Transformer model. Although DALL-E outputs were seen as astonishing, a novel category of generative models demonstrated that it was only the beginning.

The same year of the release of the DALL-E paper, \cite{ADM} showed that diffusion models could surpass GANs on class-conditional image generation. Then, GLIDE \citep{glide} introduced diffusion models for text-to-image synthesis and obtained higher-quality images than DALL-E, being trained on the same dataset. The authors of GLIDE use a Transformer language model to embed the textual image descriptions, and then diffusion models conditioned on the embeddings to produce $256 \times 256$ images. But then during the development of this thesis, two major works were released a month apart: DALL-E 2 \citep{dalle2} and Imagen \citep{imagen}. 

DALL-E 2 is very different to its former version. Similarly to GLIDE, it decodes embeddings to generate images. However, the embeddings come from CLIP \citep{clip}, a vision-language model learning image-text representations. We dedicate a section of this thesis on CLIP (\autoref{sec:clip}). So to keep it short, the two CLIP encoders produce embeddings for images and texts respectively, where the cosine similarity between two embeddings coming from the same image-caption pair is assumed to be higher than uncorrelated embeddings. DALL-E 2 leverages these pre-trained encoders and learns a prior model to translate from a CLIP text embedding to a CLIP image embedding. They also use cascaded diffusion models \citep{cascade} to upsample the images from $64 \times 64$ to $1024 \times 1024$. The structure of the method we propose in this thesis is very close to DALL-E 2, thus reading \autoref{sec:method} explaining our method can help to understand DALL-E 2 in depth.

Imagen on the other hand is more similar to GLIDE, but instead of training from scratch the language model, the authors reuse a large frozen Transformer model trained on a massive text-only corpus, called the T5 model \citep{T5}. They exhibit that increasing the size of the language model leads to higher improvements than increasing the size of the diffusion model. Besides, they double the size of the captioned images dataset and propose a few tricks to generate more realistic images, including architecture modifications and the introduction of \textit{dynamic thresholding} to improve guidance, which we therefore use in our method and detail in \autoref{sec:dynamic thresholding}. The authors claim that their model outperforms DALL-E 2 since they obtain a lower FID (metric described in \autoref{sec:experiments}) on the validation set of the MS-COCO dataset \citep{coco}. We compare these two models with our method in \autoref{sec:comparison}.

Due to their recent release, no replication of DALL-E 2 or Imagen has been fully implemented until now. The most recent and efficient replication of text-to-image generation model is DALL-E mini, inspired by DALL-E. DALL-E mini demonstrates that decent performances in text-to-image generation can still be achieved despite using smaller datasets and models. Nevertheless, DALL-E mini doesn't take into account the recent breakthroughs in image generation initiated by diffusion models. We therefore aim to complete their work by replicating a text-to-image model which considers diffusion models.

\chapter{Background}
\label{sec:background}

We introduce in this chapter all the required knowledge to understand our method. It starts with a reminder about diffusion models in \autoref{sec:diffusion_models}, as well as the different ways to guide them to generate more realistic images. In particular, we introduce a new guidance method called image guidance. We then dedicate \autoref{sec:clip} to CLIP embeddings \citep{clip}, which are extensively used by our method. We assume the reader is already familiar with basic machine learning and deep learning concepts. Otherwise, we recommend reading \cite{bishop} and \cite{deeplearning}.

\section{Diffusion Models}
\label{sec:diffusion_models}

\subsection{Introduction}
\label{sec:introduction_diffusion_models}
Diffusion models are originally based on a modelling approach of molecular systems, called Langevin dynamics. They were first introduced by \cite{sohldickstein2015}, and progressively appeared to be a serious alternative to traditional generative models such as GANs or VAEs, obtaining outstanding results in text-to-image generation \citep{glide, dalle2, imagen}.
Moreover, their generative capacity can be applied to synthesize various sources of data: image \citep{DDPM}, text \citep{diffusion_text}, speech \citep{fastdiff}, music \citep{musicdifmodel}, video \citep{videodifmodel}, or times series \citep{timegrad}.
We consider image generation here since this is the case we are interested in, but concepts are similar for other modalities. 

\begin{figure}
    \centering
    \includegraphics[width=1\linewidth]{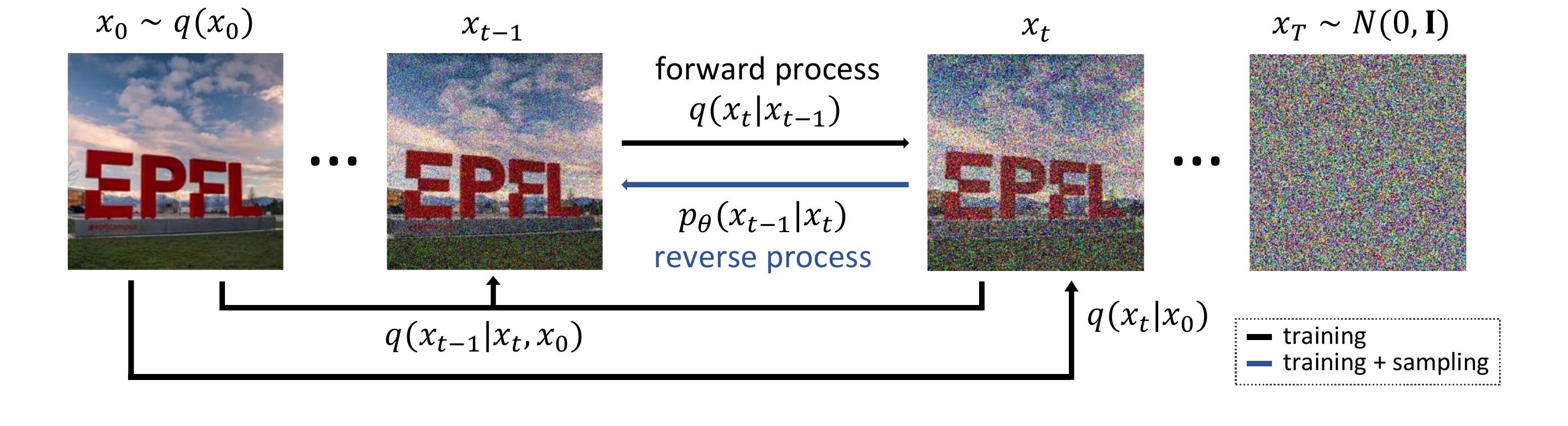}
    \caption{The forward (diffusion) process progressively adds noise to the image, whereas the reverse (inference) process removes the noise to retrieve the initial image.}
    \label{fig:diffusion}
\end{figure}

Let's suppose that we gradually add random noise to each pixel of an image. After a sufficient amount of noising steps, the image becomes itself pure noise and the signal is totally destroyed. Diffusion models try to learn the reverse process, i.e., to iterably recover the initial image from the noisy image. In mathematical terms, we call the initial image $x_0$, obtained from the data distribution $q(x_0)$. $x_T$ is the final noisy image, where $T$ is the number of noising steps sequentially applied to $x_0$. All the noising steps $x_1, \ldots, x_{T}$ are latent variables, with the same dimensionality as $x_0$. We define this progressive noising process as being the \textit{forward process} or \textit{diffusion process}, with the following distribution:
\begin{align}
    q(x_1, \ldots, x_T) := \prod^T_{t=1} q(x_t | x_{t-1}).
\end{align}
Note that we assume here that the forward process is a Markov chain, making this diffusion model a Denoising Diffusion Probabilistic Model (DDPM) \citep{DDPM}. Non-Markovian forward processes could have also been considered, leading to models such as DDIM \citep{DDIM}.
If we assume that the noise added to each step is Gaussian, then we can define:
\begin{align}
    q(x_t | x_{t-1}) := \mathcal{N}\left(x_t; \sqrt{1 - \beta_t} x_{t-1}, \beta_t \mathbf{I}\right)
\end{align}
where $\beta_1, \ldots, \beta_T \in (0, 1)$ is the \textit{variance schedule}. It controls the level of noise added at each step.
We observe that if $\beta_1 < \beta_2 < \ldots < \beta_T$, then for $T \to \infty$ the latent $x_t$ becomes an isotropic Gaussian random variable, i.e., $x_T \sim \mathcal{N}(0,\mathbf{I})$. This is a desired property since it enables one to generate samples simply by drawing from a Gaussian distribution, similarly to generative models such as GANs or VAEs. Linear \citep{DDPM} and cosine \citep{improved-DDPM} variance schedules imply this property.
By defining $\alpha_t := 1 - \beta_t$ and  $\bar{\alpha}_t := \prod^t_{s=1} \alpha_s$, the following reparametrization emerges from the forward process \citep{DDPM}:
\begin{align}
    q(x_t|x_0) &:= \mathcal{N}\left(x_t; \sqrt{\bar{\alpha}_t} x_{0}, (1 - \bar{\alpha}_t) \mathbf{I}\right)\\
    &:= \sqrt{\bar{\alpha}_t} x_{0} + \sqrt{1 - \bar{\alpha}_t} \epsilon
\end{align}
with $\epsilon \sim \mathcal{N}(0, \mathbf{I})$, since the sum of Gaussian random variables is also Gaussian. 
Thus, this marginal distribution allows us to sample any arbitrary step $x_t$ conditioned on the image $x_0$, which is handy for training the model as we will see.

We are now interested to learn the reverse distribution $q(x_{t-1}|x_{t})$, because after sampling from $p(x_T) \sim \mathcal{N}(x_T; 0,\mathbf{I})$, we could just run the process in reverse to obtain a sample of $q(x_0)$, i.e., a synthetic image. However, we cannot easily estimate $q(x_{t-1}|x_{t})$ since it requires access to the full data distribution. Instead, we approximate it by learning a model $p_\theta$ called the \textit{reverse process}, such that
\begin{align}
    p_\theta(x_0,\ldots,x_T) := p(x_T) \prod^T_{t=1 }p_\theta(x_{t-1}|x_t).
\end{align}
Besides, we define
\begin{align}
    p_\theta(x_{t-1}|x_t) := \mathcal{N}(x_{t-1}; \mu_\theta(x_t, t), \Sigma_\theta(x_t, t)).
\end{align}
where the mean $\mu_\theta(x_t, t)$ is a neural network and the variance $ \Sigma_\theta(x_t, t)$ can be computed in different ways. \cite{DDPM} fixed this variance to a constant, but \cite{improved-DDPM} showed that learning the variance was beneficial to reduce the number of diffusion steps. They parameterize the variance as the following interpolation
\begin{align}
    \Sigma_\theta(x_t, t) = \exp(v \log \beta_t + (1 - v) \log \Tilde{\beta_t})
\end{align}
with $\Tilde{\beta_t} = \beta_t\frac{1 - \bar{\alpha}_{t-1}}{1-\bar{\alpha}_t}$ and $v$ is the output of a neural network.

The combination of the forward process $q$ and the backward process $p$ can be interpreted as a variational auto-encoder \citep{VAE}, leading to the optimization of the usual variational lower bound (VLB) on negative log likelihood (also called the ELBO), defined as follows:
\begin{align}
    L_{\text{vlb}} &:= L_0 + L_1 + \ldots + L_{T-1} + L_{T} \\
    L_{0} &:= -\log p_\theta (x_0|x_1) \\
    L_{t-1} &:= D_{KL}(q(x_{t-1}|x_t, x_0) || p_\theta (x_{t-1}|x_t)) \\
    L_{T} &:= D_{KL}(q(x_{T}|x_0) || p(x_T))
\end{align}
where $D_{KL}$ is the Kullback–Leibler divergence.
Let first observe that $L_T$ is constant with respect to the parameters $\theta$ and $L_0$ can be easily evaluated using the CDF of the Gaussian distribution. For now, the only missing part is how to compute the posterior distribution $q(x_{t-1}|x_t, x_0)$. Fortunately, as exhibited by \cite{DDPM}, the posterior is tractable using Bayes theorem when it is conditioned on $x_0$:
\begin{align}
    q(x_{t-1}|x_t, x_0) &= \mathcal{N}(x_{t-1}; \Tilde{\mu}_t(x_t, x_0), \Tilde{\beta}_t\mathbf{I}) \\ 
    \Tilde{\mu}_t(x_t, x_0) &:= \frac{\sqrt{\bar{\alpha}_{t-1}}\beta_t}{1-\bar{\alpha}_{t}}x_0 + \frac{\sqrt{\alpha_t}(1-\bar{\alpha}_{t-1})}{1-\bar{\alpha}_{t}}x_t  \\
    \Tilde{\beta}_t &:= \frac{1-\bar{\alpha}_{t-1}}{1-\bar{\alpha}_{t}} \beta_t.
\end{align}
In consequence, all terms (except $L_0$) of $L_{\text{vlb}}$ are some $KL$ divergence between two Gaussian distributions, enabling us to evaluate them with the closed form expression. During training, we can apply stochastic gradient descent by considering only one random term of $L_{\text{vlb}}$.

However, \cite{DDPM} found out that optimizing a different loss could improve sample quality, but at the cost of a lower log likelihood:

\begin{align}
    L_{\text{simple}} = E_{x_0 \sim q(x_0), \epsilon \sim \mathcal{N}(0,\mathbf{I}),t \sim \mathcal{U}(\{1,\ldots,T\})}[\Vert \epsilon_\theta(\underbrace{\sqrt{\bar{\alpha}_t} x_{0} + \sqrt{1 - \bar{\alpha}_t} \epsilon}_{x_t}, t) - \epsilon\Vert^2]
\end{align}

where the predicted noise $\epsilon_\theta$ is related to the mean $\mu_\theta$ of the reverse process by the following 1-to-1 mapping:
\begin{align}
    \mu_\theta(x_t, t) = \frac{1}{\sqrt{\alpha_t}}\left(x_t - \frac{\beta_t}{\sqrt{1 - \bar{\alpha}_t}} \epsilon_\theta(x_t, t)\right).
\end{align}
$L_{\text{simple}}$ can be interpreted as the mean-squared error between the true noise added on the initial image and the noise predicted by the model with the noisy image and timestep as input. Moreover, it resembles to previous denoising score matching (\cite{score-matching}), with the score function $\nabla_{x_t} \log p(x_t) \propto \epsilon_\theta(x_t, t)$. 

However, $L_{\text{simple}}$ does not depend on the variance $\Sigma_\theta(x_t, t)$, preventing the learning of its parameters. Thus, \cite{improved-DDPM} proposed a new hybrid weighted sum objective:
\begin{align}
    L_{\text{hybrid}} = L_{\text{simple}} + \lambda L_{\text{vlb}}
\end{align}
with $\lambda$ set to $0.001$. Stop-gradient is applied to the $\epsilon_\theta$ term inside $L_{\text{vlb}}$, implying that $\mu_\theta$ is guided only by $L_{\text{simple}}$ while $\Sigma_\theta(x_t, t)$ is learnt using $L_{\text{vlb}}$. \cite{improved-DDPM} showed that $L_{\text{hybrid}}$ obtain lower log likelihood than $L_{\text{simple}}$. Hence, we use $L_{\text{hybrid}}$ for all our experiments.

\subsection{Conditioning and guidance}

When we consider image generation, we often want to specify the content of the synthetic images. Therefore, generative models have been easily adapted to take into consideration extra information to have control on the generation. While early models were only capable of producing samples of a specific class label, alignDRAW \citep{alignDRAW} demonstrated that it was possible to consider captions describing the contents of the images to produce scene compositions unseen in the dataset. However, the sample quality was low and most of the images were blurry. It is only when DALL-E \citep{dalle} came out that generative models were able to produce realistic outputs given an image caption. We now show how to take into account conditional information in the context of diffusion models.

Diffusion models have two distinct ways to integrate conditional information and we first need to explain the difference between the two. \textbf{Conditional generative models} try to learn the probability distribution $p(x|y)$, e.g., generating an image $x$ belonging to the class $y$ or matching a caption $y$. Applied to diffusion models, it simply consists of learning the conditional model $\epsilon_{\theta}(x_t, t|y)$ instead of the unconditional model $\epsilon_{\theta}(x_t, t)$. Thus, the model directly takes $y$ as input to condition the generation of $x$ during both training and sampling. How $y$ is concretely incorporated into the model is explained in \autoref{sec:diffusion_architecture}.

On the other hand, \textbf{guidance methods} don't change the model structure, and are only used during sampling. Guidance slightly modifies the output of the model at each diffusion step to help the generative process to go in the desired direction. This small update often takes the form of a gradient ascent step and the desired direction can again be a condition $y$ such as a class label or a caption. \cite{ADM} exhibited that guidance could greatly improve sample quality.
Finally, guidance methods and conditional generative models are complementary and can therefore be used jointly. Some guidance approaches such as classifier-free guidance even require a conditional model. We now review the different guidance methods usey by diffusion models.

\subsubsection{Classifier guidance}

Introduced by \cite{ADM}, classifier guidance leverages an extra image classifier model $p_\phi(y|x_t, t)$, which already conveys some knowledge of the dataset, to help the generating process. The gradient $\nabla_{x_t} \log p_\phi(y|x_t, t)$ is exploited to guide the sampling process in the direction of class label $y$:
\begin{align*}
    \Tilde{\mu}_\theta(x_t, t|y) = \mu_{\theta}(x_t, t) + w \Sigma_\theta (x_t, t) \nabla_{x_t} \log p_\phi(y|x_t, t),
\end{align*}
with $w \geq 0$ being the classifier guidance scale.
Note that $p_\phi(y|x_t, t)$ considers noisy images $x_t$, which therefore requires to train a noise-aware classifier.

This notion of classifier guidance can be extended to other sorts of models, and not only classifiers. For instance, the authors of GLIDE \citep{glide} replace the image classifier by a CLIP model \citep{clip} to help generate images from text. CLIP provides a measure of similarity between an image and a caption (we explain more in detail how CLIP works in the next section, \autoref{sec:clip}) and taking the gradient with respect to the image enables to guide the sampling process in the direction of the caption. However, the authors observe that classifier-free guidance performs better than CLIP guidance.

\subsubsection{Classifier-free guidance}
\label{sec:classifier_free}

Depending on a separate model is inconvenient and it complicates the training pipeline. Moreover, for classifier guidance, the number of classes is limited, preventing us from conditioning on complex information to generate more elaborated image compositions. Thus, \cite{classifierFree} proposed classifier-free guidance, which only relies on a single diffusion model. Classifier-free guidance considers a conditional diffusion model $\epsilon_{\theta}(x_t, t|y)$ that can be made unconditional by replacing occasionally during training the condition $y$ by an empty condition $\emptyset$, e.g., by setting the caption to an empty string. The model $\epsilon_{\theta}(x_t, t|\emptyset)$ can then be used to generate unconditional images. Classifier-free guidance therefore consists in updating the model output using a linear combination between $\epsilon_{\theta}(x_t, t|\emptyset)$ and $\epsilon_{\theta}(x_t, t|y)$ in the following way:
\begin{align*}
    \Tilde{\epsilon}_\theta(x_t, t|y) = \epsilon_{\theta}(x_t, t|\emptyset) + s \cdot (\epsilon_{\theta}(x_t, t|y) - \epsilon_{\theta}(x_t, t|\emptyset)),
\end{align*}
with $s \geq 1$ being the guidance scale. This update can be understood as an attempt to move further in the direction of the conditional model, while moving away from the unconditional model. It is performed at each diffusion step when sampling and \cite{classifierFree} have shown that it improves sample quality.

\subsubsection{Image guidance}
\label{sec:image_guidance}

We propose a new guidance method which has the potential to perform image inpainting and image editing. Image inpainting is a task which aims at reconstructing missing parts in an image, whereas image editing allows the modification of some elements of an image. We call this novel approach image guidance, and it is inspired by other guidance methods. Instead of using a gradient or an implicit gradient, we consider directly an extra image $z$ which will guide the sampling process. At each sampling step, we move in the direction of $z$ by updating $x_t$ in the following way:
\begin{align*}
    \Tilde{x}_t = x_t + w \cdot d_t \cdot (z - x_t),
\end{align*}
where $w \geq 0$ is the image guidance scale and $d_t$ is a decay depending on the timestep. We consider a linear decay, i.e., $d_t = t/T$, but other forms of decay could be considered. The decay helps to make the base image $z$ contribute more to the output at the first steps of the reverse process.

This approach has the benefit as other guidance methods to be applied only during sampling and not during training. It therefore contrasts with other methods including GLIDE \citep{glide} and Palette \citep{palette} which need to fine-tune and modify the model architecture if they want to perform image inpainting and editing. It is possible to combine image guidance with conditioning to fill the missing regions of an image and to edit the image in a desired way by providing an image or a text embedding as conditioning. We show in \autoref{sec:results_image_guidance} what image guidance is able to do.

\subsubsection{Influence of the guidance scale and dynamic thresholding}
\label{sec:dynamic thresholding}

For each guidance method, we scale the level of guidance by a constant factor, i.e., the guidance scale. This parameter is very important and its impact on the sampling process is different for each type of guidance method. For image guidance, the guidance scale is highly sensitive because it moves at each step the image in the same direction. Its impact is then straightforward; a larger image guidance scale implies to sample an image closer to the image used for guidance. The right combination between the image guidance scale and its decay is hence crucial to obtain the desired samples. For classifier guidance and especially when the diffusion model is conditioned on a class label, the guidance scale represents a trade-off between diversity and sample fidelity as exhibited by \cite{DDPM}, with a higher diversity and lower sample fidelity when the scale is small and vice versa if the scale is large. When we condition on captions, large guidance scales also lead to more accurate text-image alignments. However, it further causes a train-test mismatch which engenders over-saturated and unnatural images, as exposed by the authors of Imagen \citep{imagen}. Since the guidance scale is large, the values of the latent $x_t$ can exceed the bounds of the training data, i.e., the range $[-1,1]$, for any timestep $t$ during sampling.

To mitigate this issue, \cite{DDPM} introduced \textit{static thresholding}, which simply consists of clipping the pixel values of every latent variable $x_t$ to $[-1,1]$. Nevertheless, \cite{imagen} have shown that the effect of static thresholding is moderate and that therefore the generated images still suffer from saturation. For that reason, they propose \textit{dynamic thresholding}, a method which actively tries to push pixel values which are close to saturation, i.e., in the vicinity of $-1$ or $1$, towards lower absolute values. At each sampling step $t$, they compute the prediction of $x_0$ as 
\begin{align*}
    \hat{x}_0^t = \sqrt{\frac{1}{\bar{\alpha}_t}} x_t - \sqrt{\frac{1}{\bar{\alpha}_t} - 1} \cdot \epsilon_\theta(x_t, t).
\end{align*}
They then consider the $99.5$ percentile absolute pixel value in $\hat{x}_0^t$ and call it $s$. Next if $s > 1$, they clip $x_0$ to the interval $[-1,1]$ and then divide by $s$. This procedure allows them to increase the guidance scale to obtain better text-image alignments, while keeping good image quality. Besides, we empirically noticed that dynamic thresholding is effective to prevent image guidance from saturating the image.

\section{CLIP embeddings}
\label{sec:clip}

Finally, we end this chapter by reviewing CLIP \citep{clip}, an efficient method to learn image representation using natural language supervision. Recent works have exhibited that scaling up the size of the dataset with data scraped from the internet could lead to significant model improvements. In particular, GPT-like \citep{GPT3} and BERT-like \citep{Roberta} models have demonstrated that a large amount of texts coupled with an efficient self-supervised learning approach was required for natural language understanding. In contrast, computer vision has long been based on pure supervision using the so-called "gold labels", such as distinct class labels. These annotations are often crowd-sourced and are therefore difficult to obtain in a sufficient amount. 

\begin{wrapfigure}{l}{0.5\textwidth}
    \centering
    \includegraphics[width=1\linewidth]{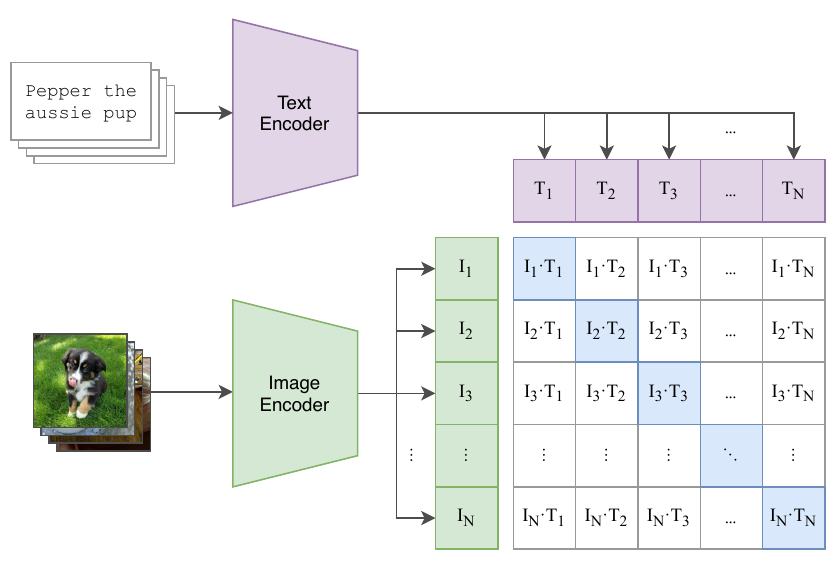}
    \caption{CLIP training. It aims to maximize the dot product between embeddings obtained from similar captions and images. Figure borrowed from \cite{clip}.}
    \label{fig:clip}
\end{wrapfigure}

Based on these observations, CLIP leverages the large quantity of natural texts accompanying images on the internet to scale up the size of its dataset. The latter consists of 400,000 pairs of image and caption, where the caption is supposed to encapsulate the semantic content of the image. CLIP uses then two models (text and image encoders)
to produce a text embedding and an image embedding respectively. The architecture of the text encoder is a Transformer model \citep{transformer}, whereas the one of the image encoder is a Vision Transformer \citep{vit}.
The two encoders are jointly trained in a contrastive way, by maximizing the cosine similarity between two embeddings of caption-image pairs while minimizing it for non-associated caption and images (see \autoref{fig:clip}). For this reason, the acronym of CLIP stands for \textit{Contrastive Language-Image Pre-training}. This whole procedure hence implies that the image and text embeddings belong to the same multimodal latent space and that similar captions and images should be close in the cosine similarity sense.

Moreover, CLIP embeddings have some useful properties.
\cite{clip} exhibited their zero-shot abilities to perform well on out-of-distribution samples, being robust to natural distribution shifts. Indeed, benefiting from pre-training on a massive amount of various samples, CLIP embeddings along with a simple linear classifier obtain better performances on different distribution shifts dataset including ImageNetV2 \citep{imagenetV2} and ImageNet Adversarial \citep{imagenet_adversarial} than the ResNet101 model \citep{resnet}, without seeing any sample of ImageNet \citep{imagenet}. Besides, the embeddings can be easily transferred to downstream tasks. \cite{CLIPanalysis} were able to match the state-of-the-art on many vision and language tasks such as Visual Question Answering (VQA) using the CLIP embeddings.
These properties enable us to leverage the CLIP encoders on any image or text dataset to obtain linked visual and textual representations. We show how we use these CLIP image and text embeddings to help generate images in the next chapter.

\chapter{Method}
\label{sec:method}

\section{Overview}

\begin{figure}[h]
    \centering
    \includegraphics[width=1\linewidth]{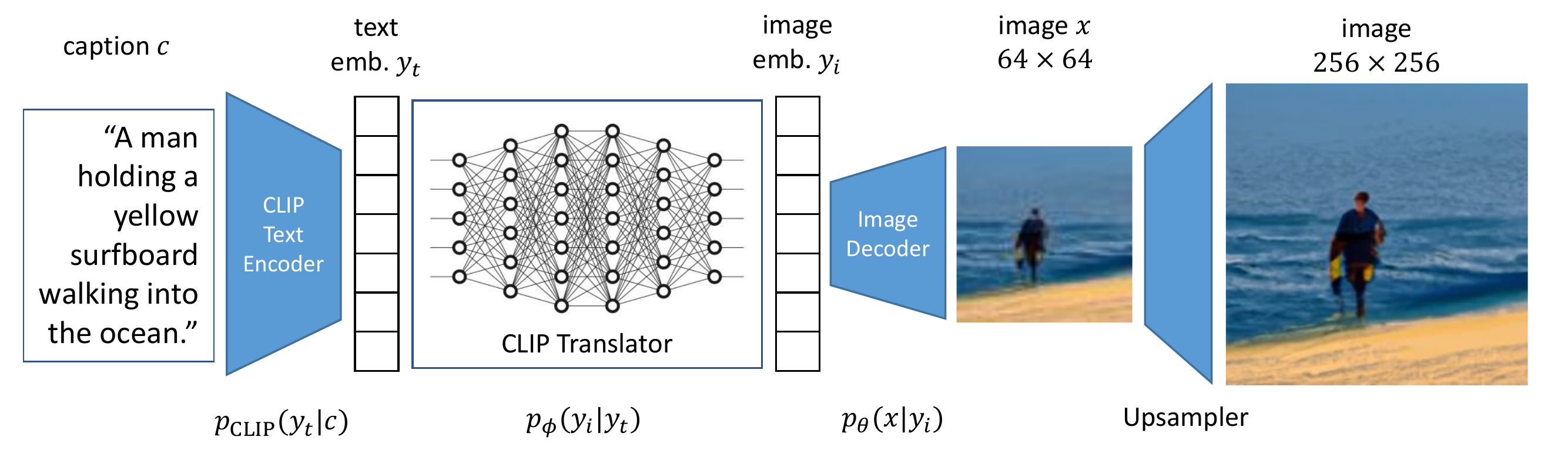}
    \caption{Our approach to synthesize images from text. Starting from an image description, we sequentially apply different models to obtain an image of resolution $256 \times 256$. It is similar to DALL-E 2, but the models considered are different.}
    \label{fig:pipeline}
\end{figure}

We describe in this section the method we use to generate images from text. It is predominantly inspired by DALL-E 2 \citep{dalle2}, but with slight modifications. In particular, it uses CLIP embeddings to represent texts and images with the aim of transferring from one modality to another. It starts by considering as input a text caption $c$ describing the content of the desired image. The text of this caption is then encoded by a CLIP encoder into a text embedding $y_t$ of length $512$. Next, we use a model called the \textit{CLIP translator} to translate the text embedding $y_t$ into an image embedding $y_i$, also of length $512$. We finally employ a diffusion model named the \textit{image decoder} to obtain an image $x$ of resolution $64 \times 64$ from the image embedding. Optionally, we can upsample our image to resolution $256 \times 256$ by exploiting a super-resolution model.
\autoref{fig:pipeline} illustrates the full pipeline, representing how the different models fit into each other. Note that it only depicts how we sample images; the training of this pipeline is done differently and independently for each model as we will see in the next sections.

Our method handles then several models with different purposes:
\begin{itemize}
    \item \textbf{Image decoder:} This is our main model and the most important one. The image decoder is a diffusion model which generates image $x$, conditioned on a CLIP image embedding $y_i$, i.e., it models the probability $p_\theta(x| y_i)$. We describe in detail the model as well as its architecture and its training in the next \autoref{sec:image_decoder}.
    \item \textbf{CLIP translator:} This model translates CLIP text embeddings $y_t$ into CLIP image embeddings $y_i$. Despite the fact that text and image embeddings are already supposed to be close for similar contents as we have seen in \autoref{sec:clip}, the CLIP translator learns to reduce the differences between the two. The model can be then written in probabilistic terms as $p_\phi(y_i|y_t)$. As for the image decoder, we dedicate a section to this model (\autoref{sec:clip_translator}).
    \item \textbf{CLIP encoders:} We use trained CLIP encoders\footnote{available at \url{https://github.com/openai/CLIP}} to embed caption $c$ and image $x$. The CLIP text encoder $p_{\text{CLIP}}(y_t|c)$ is a Transformer \citep{transformer} with 63M parameters and the CLIP image encoder $p_{\text{CLIP}}(y_i|x)$ is a ViT-B/32 \citep{vit} with 86M parameters. The image encoder is only used during the trainings of the image decoder and the CLIP translator, to obtain the embeddings of the images.
    \item \textbf{Super-resolution model:} The image decoder only creates images of resolution $64 \times 64$, which is low. Thus, we increase the resolution of our generated images to $256 \times 256$, upsampling with a super-resolution model \citep{sr3}. This upsampler model\footnote{available at \url{https://github.com/openai/guided-diffusion}} consists of a diffusion model trained on ImageNet \citep{imagenet}, being therefore restricted to ImageNet-like pictures, e.g., the upsampler doesn't recognize and doesn't produce natural text or numbers on images well. 
    
    Super-resolution models play a crucial role in DALL-E 2 and Imagen, allowing to obtain detailed $1024 \times 1024$ images using cascaded diffusion models \citep{cascade}. However, training from scratch one or two super-resolution models is complex and requires images of larger resolution, which considerably increases the size of the dataset and involves heavier computations. Being limited in the amount of available computational resources, we opt therefore for an already trained upsampler model even if it is limited.
\end{itemize}

Then generating images $x$ from a caption $c$ can be described probabilistically by the following equation
\begin{align*}
    p_{\theta,\phi,\text{CLIP}}(x|c) =  p_{\theta}(x|y_i) p_{\phi}(y_i|y_t) p_{\text{CLIP}}(y_t|c).
\end{align*}
We keep the upsampler out of this equation since it is used optionally.
Note that since we reuse the CLIP encoders and the upsampler model, we only need to train the image decoder and the CLIP translator, i.e., finding parameters $\theta$ and $\phi$. Moreover, they can be trained independently and with different datasets, as they consider different types of data. We hence dedicate the next two sections to them.

\section{Image Decoder}
\label{sec:image_decoder}

\begin{algorithm}
\caption{Image decoder (diffusion model) sampling}
\label{alg:sampling}
\begin{algorithmic}[1]
\Require diffusion model $(\epsilon_\theta, \Sigma_{\theta})$, image embedding $y_i$, guidance scale $s$
\State $x_T \sim \mathcal{N}(0, \mathbf{I})$
\For{$t = T,\ldots,1$}
\State $\epsilon = \epsilon_{\theta}(x_t, t|\emptyset) + s \cdot \left(\epsilon_{\theta}(x_t, t|y_i) - \epsilon_{\theta}(x_t, t|\emptyset)\right)$ \Comment{Apply classifier-free guidance}
\State $\Tilde{\epsilon} = \text{dynamic\_thresholding}(\epsilon)$ \Comment{Apply dynamic thresholding}
\State $\mu_\theta(x_t, t) = \frac{1}{\sqrt{\alpha_t}}\left(x_t - \frac{\beta_t}{\sqrt{1 - \bar{\alpha}_t}} \Tilde{\epsilon} \right)$
\State $z \sim \mathcal{N}(0, \mathbf{I}) \text{ if } t > 1 \text{, else } z = 0 $
\State $x_{t-1} = \mu_\theta(x_t, t) + \Sigma_{\theta}(x_t,t) \odot z$
\EndFor \\
\Return $x_0$
\end{algorithmic}
\end{algorithm}

We focus on this section on the image decoder, which generates images conditioned on a CLIP image embedding. The image decoder is a diffusion model, in particular a DDPM \citep{DDPM}. It allows us to use classifier-free guidance (described in \autoref{sec:classifier_free}) along with dynamic thresholding to sample from it. Algorithm \ref{alg:sampling} describes the sampling process to generate images. How dynamic thresholding is applied is also explained in detail in \autoref{sec:dynamic thresholding}. We set the guidance scale $s$ to $6$ and use the $99.5$ percentile for dynamic thresholding. Note that since the generating process is non-deterministic, the same image embedding can engender different variations of the same image content.

We first introduce in the next subsection the model architecture of the diffusion model. We then describe the dataset used to train the diffusion model as well as the training procedure.

\subsection{Architecture}
\label{sec:diffusion_architecture}

\begin{figure}[h]
    \centering
    \includegraphics[width=1\linewidth]{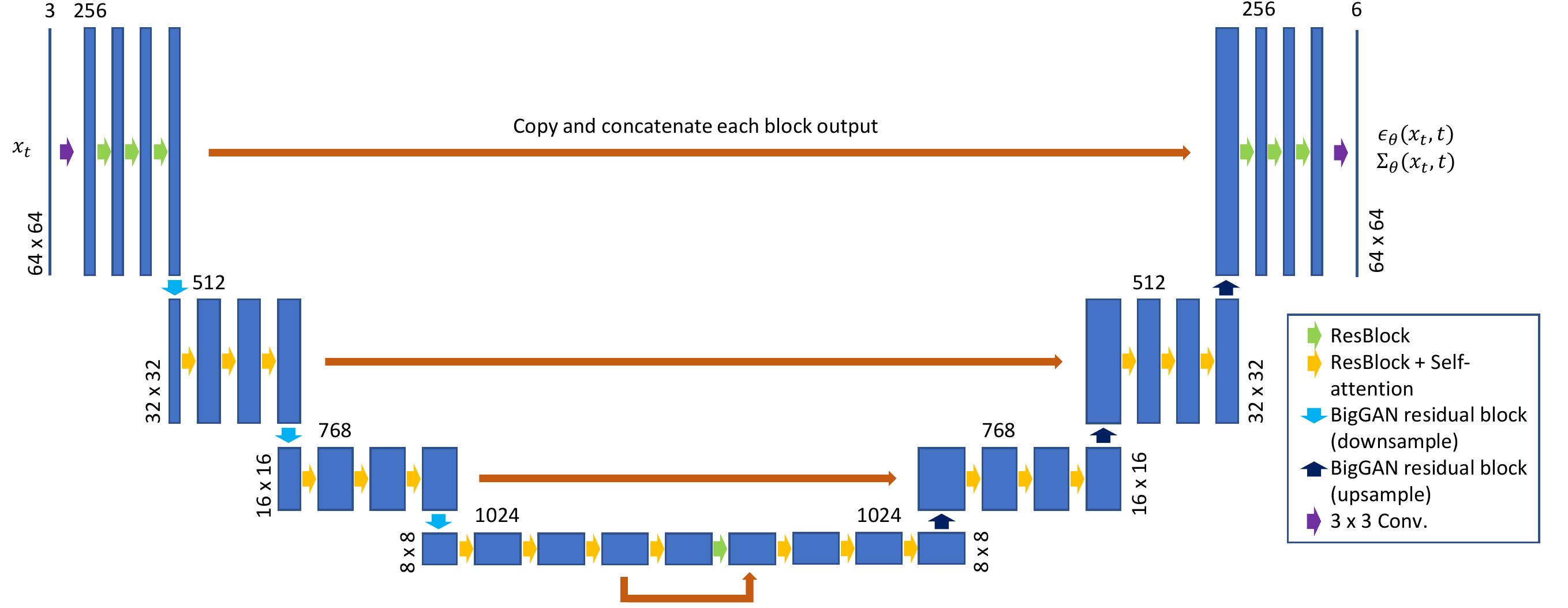}
    \caption{The U-Net architecture of the diffusion model (image decoder). It processes a RGB $64 \times 64$ image $x_t$ along with a timestep $t$ and outputs the noise $\epsilon_\theta$ and the variance $\Sigma_\theta$.}
    \label{fig:diffusion_model}
\end{figure}

\begin{wrapfigure}{r}{0.5\textwidth}
    \centering
    \includegraphics[width=1\linewidth]{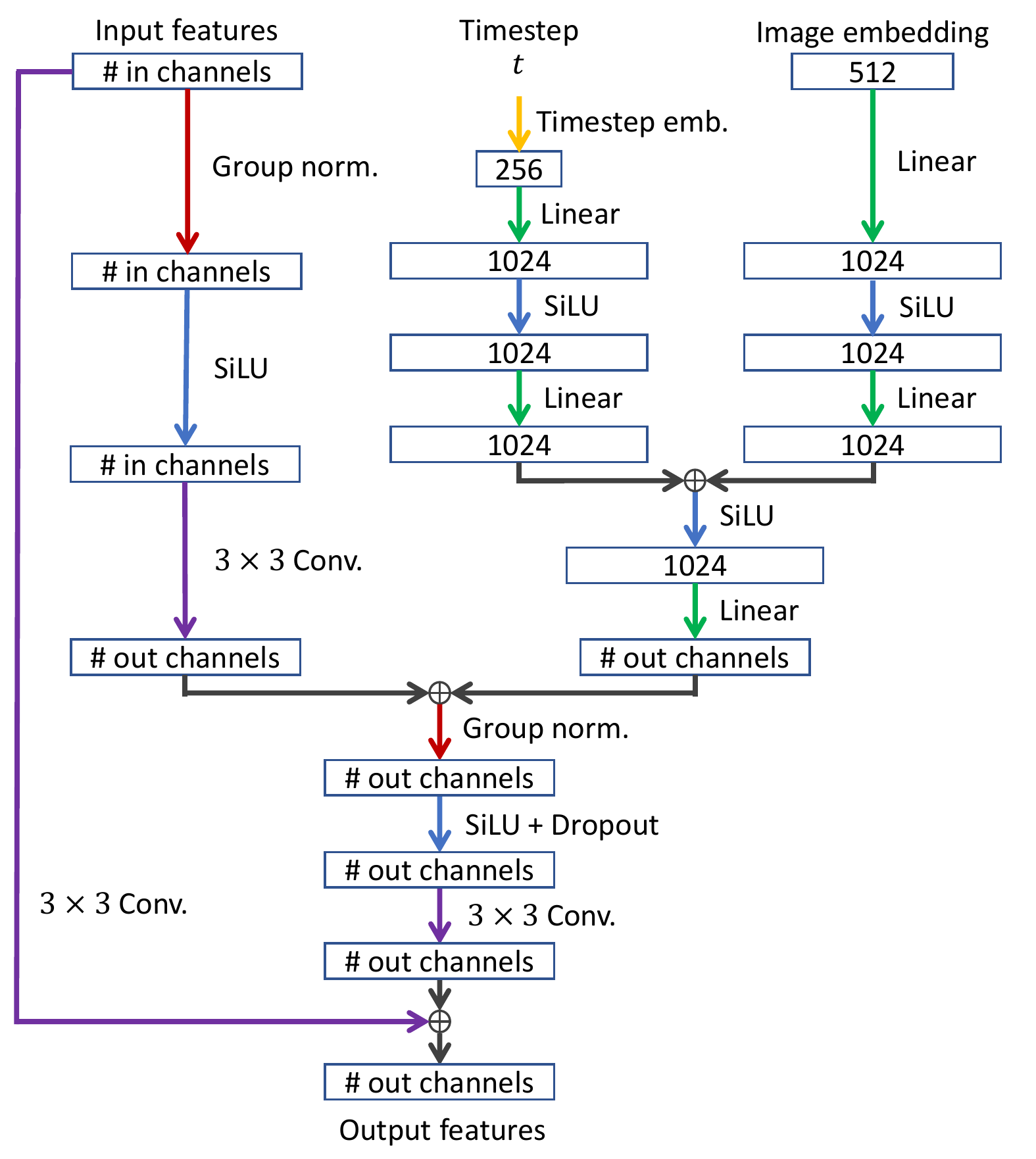}
    \caption{Structure of the ResBlock. The BigGAN residual blocks are very similar, but with an extra upsample/downsample operation to modify the spatial dimension size.}
    \label{fig:resblock}
\end{wrapfigure}

We opt for a U-Net architecture \citep{unet} as proposed by \cite{DDPM}. A U-Net aims to output a sample of the same shape of the input and was used originally for biomedical image segmentation. The architecture is composed of two distinct parts: an \textit{encoder} decreasing the resolution and increasing the number of channels, and a \textit{decoder} doing the opposite to retrieve the original shape. The encoder and decoder each contain a few resolution layers, corresponding to different spatial sizes, i.e., resolutions. The layers of the encoder and decoder with the same spatial size, are linked using skip connections, where each block output in the encoder is concatenated to the corresponding block input in the decoder. The different types of blocks in the U-Net are the following:
\begin{itemize}
    \item \textbf{ResBlock:} This is a standard resnet block \citep{resnet}, containing two convolutional layers with group normalization \citep{gn} and SiLU \citep{silu} applied before each of them, as well as a Dropout module \citep{dropout}, and a skip connection. The convolutional layers use kernels of size $3 \times 3$, a stride of $1$, and a padding of $1$. The ResBlock is also used to increase or decrease the number of channels. All the operations performed in a ResBlock are illustrated in \autoref{fig:resblock}.
    \item \textbf{Self-attention block:} Self-attention \citep{transformer} is a powerful mechanism to draw global dependencies between different parts of the input.
    The authors of Palette \citep{palette} exhibited the importance of self-attention blocks to improve sample quality, whereas \cite{cordonnier} demonstrated their high expressiveness, similar and often even larger than convolutional layers.
    Thus, we add self-attention blocks after ResBlocks, but only at low resolution levels since the computational complexity grows quadratically with the resolution.
    The spatial shape of the input is first flattened before applying self-attention, and residual connections rescaled by $\frac{1}{\sqrt{2}}$ are used to connect the input to the output.
    The activations are normalized using group normalization.
    \item \textbf{BigGAN residual block:} Introduced by \cite{BigGAN}, BigGANs use residual blocks for upsampling and downsampling the activations. \cite{ADM} found them to be beneficial to improve performance, and therefore use them to increase or decrease the resolution. They are very similar to ResBlocks, but with an upsample or downsample operation interleaved between the first group normalization and the first convolutional layer.
    \item \textbf{Timestep embedding:} The timestep $t$ is incorporated into a sinusoidal timestep embedding \citep{transformer}. This embedding is then linearly projected and integrated into each block of the U-Net, i.e., the ResBlocks, the self-attention blocks and the BigGAN residual blocks.
    \item \textbf{Conditional image embedding:} We condition the diffusion model on an image embedding. When we want to use the model unconditionally, we just set the embedding to the null vector. The image embedding is first linearly projected before being added to the timestep embedding.
\end{itemize}

Our U-Net takes as input a noisy image $x_t$ at $64 \times 64$ resolution, as well as the timestep $t$ indicating the noise level of the image.
It outputs the mean $\mu_\theta$ and the variance $\Sigma_\theta$,
which are used to generate $x_{t-1} \sim \mathcal{N}(\mu_\theta, \Sigma_\theta)$.
We draw the noise $x_T \sim \mathcal{N}(0,\mathbf{I})$ and apply recursively the U-Net to obtain a sample $x_0$.
Sampling a single image requires then $T$ forward passes of the U-Net. The resolution of the generated images is also $64 \times 64$.

The values of the hyperparameters are based on the work of \cite{ADM} with some adjustments recommended by \cite{dalle2}. The diffusion model uses $T=1000$ diffusion steps with noise being applied following a cosine schedule. The encoder and decoder of the U-Net are both composed of $4$ resolution layers, with resolution varying between $64, 32, 16,$ and $8$, and the number of channels between $256, 512, 768,$ and $1024$.
Each of these resolution layers are composed of $3$ ResBlocks, with two extra blocks at the lowest resolution level to connect the encoder and decoder parts. Self-attention is only applied at resolution layers $32, 16,$ and $8$, with $64$ channels per head. Finally, we reduce the interdependence among neurons by applying Dropout \citep{dropout} after ResBlocks with probability $0.1$.

\subsection{Data}
\label{sec:diffusion_dataset}

We train the diffusion model by jointly considering three different datasets. First, we use the \textbf{ImageNet} Large Scale Visual Recognition Challenge (ILSVRC) dataset, a subset of 1,281,167 samples of ImageNet \citep{imagenet}. It contains square images of various types of objects, animals, vehicles, people, or food, which are divided into 1000 classes. The annotations were obtained via crowdsourcing on the web. Our approach however is self-supervised, and therefore doesn't need the class labels. This dataset helps our diffusion model to generate images focused on a single item or entity.

On the other hand, the image diversity of ImageNet is limited, with objects being often out of their natural context. Besides, we would like to be able to generate diverse scenes, with different concept compositions. We thus use \textbf{CC3M} \citep{cc3m} and \textbf{CC12M} \citep{cc12m} to add more complex situations in our image dataset. These two datasets contain 2,064,294 and 8,856,123 caption-image pairs respectively, harvested from the internet. In particular, the datasets authors implemented an automatic pipeline which extracts, filters, and modifies the raw description obtained from the Alt-text HTML attribute of images on the web. They then obtain what they call conceptual captions whose specificities such as proper nouns, numbers, and dates have been removed or been substituted by hypernyms. CC12M is just a higher recall dataset than CC3M, by using less restrictive filters. Nevertheless, the captions are not necessary to train the diffusion model, since we use the CLIP image encoder to obtain directly the image embeddings. But they are for training the CLIP translator as we will see in \autoref{sec:mlp_dataset}.

The diffusion model is then trained using a total of 12,201,584 images, which are all resized to a $64 \times 64$ resolution. For rectangular images from CC3M and CC12M, white borders are added.
The CLIP image embeddings are nevertheless obtained on the original image resolution, to encapsulate more fine details. They are also standardized before they are inputted into the model.
Finally, we use data augmentation by randomly horizontally flipping half of the images as suggested by \cite{dataaugmentation}, and we scale the values of the pixels to values between $-1$ and $1$.

\subsection{Training}

\begin{algorithm}[h]
\caption{Image decoder (diffusion model) training}\label{alg:training}
\begin{algorithmic}[1]
\Require image dataset $q(x_0)$, embedding drop probability $p$
\State Initialize $\epsilon_\theta, \Sigma_\theta$
\While{ $\epsilon_\theta, \Sigma_\theta$ have not converged}
\State $x_0 \sim q(x_0)$
\State $t \sim \mathcal{U}(\{1,\ldots,T\})$
\State $\epsilon \sim \mathcal{N}(0, 1)$
\State $u \sim \mathcal{U}(0,1)$
\State $y_i = \text{CLIP}(x_0) \text{ if } u \geq p \text{, else } y_i = \vec{0} $ \Comment{Randomly drop the CLIP image embedding}
\State $x_t = \sqrt{\bar{\alpha}_t} x_{0} + \sqrt{1 - \bar{\alpha}_t} \epsilon$
\State Take gradient descent step on
    $\nabla_\theta \left(\Vert \epsilon_\theta(x_t, t | y_i) - \epsilon\Vert^2 + \lambda \cdot \text{sg$_{\epsilon_\theta}$}[L_t]\right)$  \Comment{Optimize $L_{\text{hybrid}}$}
\EndWhile \\
\Return $\epsilon_\theta, \Sigma_\theta$
\end{algorithmic}
\end{algorithm}

The diffusion model is then trained by minimizing the hybrid loss $L_{\text{hybrid}}$ (see \autoref{sec:introduction_diffusion_models}) using gradient descent and the backpropagation algorithm \citep{backpropagation}. The training procedure is described by Algorithm \autoref{alg:training}. The abbreviation sg$_{\epsilon_\theta}$ stands for stop-gradient, avoiding the backpropagation of $L_t$ to update the parameters $\epsilon_\theta$.
Moreover, we use the Adam optimization algorithm \citep{adam} with default parameters values, i.e., $\beta_1=0.9$, $\beta_2=0.999$, and $\epsilon=1$e-$8$, but without weight decay. The learning rate is initially set to $3$e-$4$ and is then annealed at each iteration with a linear decay. We noticed some exploding gradients during training, we hence apply gradient clipping to keep the norm of the gradients reasonable. In order to make the model fit into the memory, we fix the batch size to $16$. We also maintain the exponential moving average (EMA) of the weights, and sample with the EMA model. The EMA rate is $0.9999$. Besides, we exploit the CLIP image encoder to obtain the image embedding $y_i$, but we set it to the null vector with probability $p=0.2$ for the purpose of enabling the model to generate images unconditionally and to apply classifier-free guidance. Finally, we perform 500,000 iterations (weight updates), corresponding to a bit less than 1 epoch. More iterations should still improve our model, but our computational resources are limited and the 500,000 iterations already last $8$ days on a NVIDIA Tesla V100 SXM2 32GB.

\section{CLIP Translator}
\label{sec:clip_translator}

In this section, we delve into the implementation of the CLIP translator which as its name suggests translates embeddings from one modality to another, in this case from text to image. In a similar way as it was done for the section about the image decoder, we introduce in the next subsections the model architecture, the datasets used, and the training process.

\subsection{Architecture}

Authors of DALL-E 2 propose two different architectures for their CLIP translator (which they call the prior model in their research paper). They consider either a Transformer \citep{transformer} or a diffusion model, but they demonstrate that the diffusion model generates higher-quality samples than the Transformer. We observe however that the CLIP translator simply needs to perform a vector-to-vector transformation, where the two vectors (embeddings) are already very similar because of how CLIP embeddings are produced (see \autoref{sec:clip}). Moreover, the elements of the vectors can be permuted without losing any information, since they are outputs of Transformer models. It involves that we cannot take advantage of an inductive bias such as a spatial one as it is done by convolutional layers or a sequential one for autoregressive models.

\begin{figure}
    \centering
    \includegraphics[width=1\linewidth]{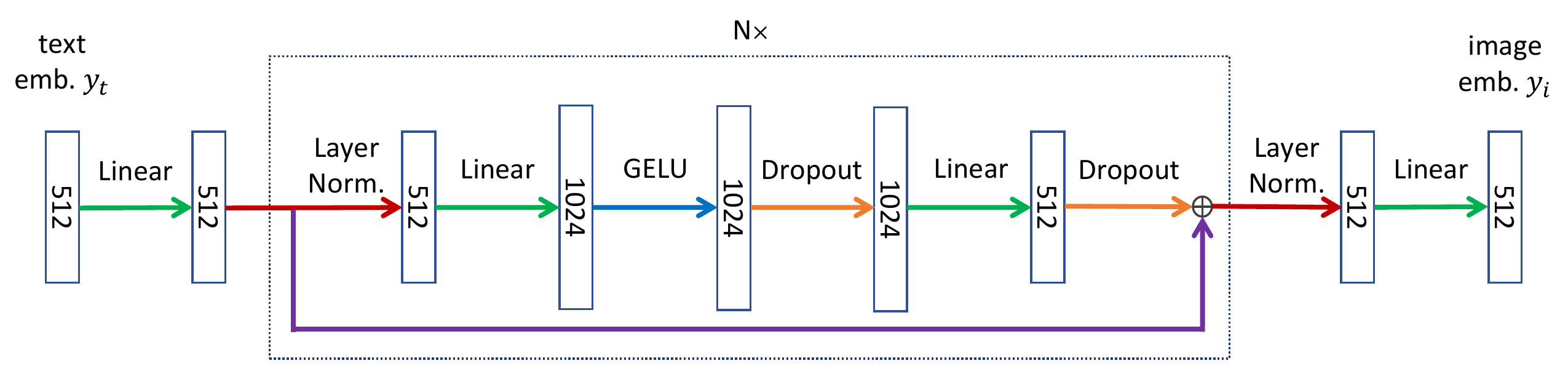}
    \caption{The multilayer perceptron architecture of the CLIP translator.}
    \label{fig:MLP}
\end{figure}

We therefore simply consider as the architecture of the CLIP translator a multilayer perceptron (MLP). \cite{MLP-Mixer} have shown that slight changes to the original MLP architecture could make this model able to compete with recent Transformer and CNN models. Besides, MLP models are incredibly easy to implement, relying often only on a few lines of codes using popular deep learning frameworks. To benefit from the extensive advancements of the past decade in deep learning, we hence integrate many architectural components of the state-of-the-art MLP-Mixer \citep{MLP-Mixer}, including layer normalization \citep{layernormalization}, Dropout \citep{dropout}, skip-connections \citep{resnet}, and GELU \citep{gelu}. The architecture is depicted in \autoref{fig:MLP}. It is mainly composed of a stack of $N$ identical layers, performing each of the operations listed above, as well as two linear projections. Dropout is used with probability $0.1$.

\subsection{Data}
\label{sec:mlp_dataset}

We use the images and captions of CC3M and CC12M (both described in \autoref{sec:diffusion_dataset}) along with the CLIP encoders to obtain pairs of image and text embeddings. It becomes a simple supervision learning task where we try to model the transformation from a caption embedding to its associated image embedding. Note that we don't use ImageNet to train the CLIP translator since it doesn't contain captions. We thought about adding the 82,783 captioned images of the training set of the MS-COCO dataset \citep{coco} to our dataset collection, but we finally decided to use MS-COCO only for testing the whole pipeline, with the aim of zero-shot learning this dataset (see \autoref{sec:testing}). 

Similarly as for the image decoder, the image embeddings are computed on the original image resolution. Both image and text embeddings are standardized, i.e., subtracted by the mean and divided by the standard deviation. Finally, we split the 10,920,397 embedding pairs into a training set (95\%) and a validation set (5\%).

\subsection{Training}

To train the CLIP translator, we simply consider the Mean Squared Error (MSE) between the image embeddings ground-truth $y_i$ and the image embeddings $\hat{y}_i$ predicted by the MLP, i.e., $L_\text{MSE} = \frac{1}{n}\Sigma^n_{k=1}((y_i)_k - (\hat{y}_i)_k)^2$. We then seek to minimize this loss function using gradient descent with AdamW \citep{adamw}, since we apply weight decay with rate $0.0001$. This is employed jointly with Dropout \citep{dropout} to avoid overfitting. The probability of Dropout is $0.1$ and the parameters of AdamW are again the default ones: $ \beta_1 = 0.9$, $\beta_2 = 0.999$, $\epsilon=1$e-$8$, and the learning rate is equal to $1$e-$3$. Finally, we consider a batch size of $256$.

The hyperparameter values above are common to train machine learning models. We decide to not spend too much time tuning them, because it only marginally improves the performance of the CLIP translator. Instead, we focus on the number of layers $N$ of the MLP which empirically seems to be the largest source of variations of the loss function. Specifically, we perform grid search on its value using the validation set and we find that $N=30$ is the most effective.

Finally, we train the CLIP translator with $30$ layers for $6$ epochs and apply early stopping. The model with the lower validation loss is achieved at epoch $5$ and is kept for our pipeline. We can compare our model to an \textit{identity model} which simply outputs its input $y_t$. The identity model gets a loss of $1.5
$, while our CLIP translator achieves a loss of $0.66$, which is more than twice as small.

\chapter{Experiments}
\label{sec:experiments}

Now that we have a fully operational method to generate images from text, we can experiment its effectiveness and how well each part of the pipeline is performing. We start by simply evaluating the method to generate images of good quality: firstly in considering the whole system, and secondly in performing ablation studies where some components are removed to investigate their individual contribution.
Next, as we extensively use the CLIP embeddings, we would like to understand to which degree they can capture semantic regularities in texts and in images.
Finally, we analyze our novel image guidance method to determine what it is capable of doing. 

\section{Testing our method}
\label{sec:testing}

Before assessing the performances of our method and its components, we introduce the main metric that we will use to do so. It consists of the \textit{Fréchet Inception Distance} (FID), a common criterion to evaluate the performances of image generative models. Introduced by \cite{FID}, it aims to overcome the shortcomings of the \textit{Inception Score} (IS) from \cite{inception_score}. Even if both metrics utilize an extra image classifier, an Inception-v3 trained on ImageNet \citep{InceptionV3}, the IS only considers the distribution of the generated images to determine their quality. On the other hand, the FID exploits and compares statistics of both the generated and real-world images, the latter belonging to the same distribution of the model training set. It makes the FID correlating more to human judgement of sample quality than the IS.

In particular, the FID uses the activations produced by the last pooling layer of the Inception-v3 classifier, yielding a 2,048 feature vector for each image (real and synthetic). Next, it computes the first two moments of the activations of the real and the generated images separately, i.e., the two means $\mu_r,\mu_g$ and the two covariance matrices $\Sigma_r,\Sigma_g$ respectively. The FID is then given by
\begin{align*}
    \text{FID} = \Vert\mu_r -\mu_g \Vert^2 + \text{tr}(\Sigma_r + \Sigma_g -2(\Sigma_r \Sigma_g)^{1/2}).
\end{align*}
Lower FIDs involve in general higher image quality. We hence compute and analyze this metric for different experiments. For more exhaustivity, we also include the IS, and the improved precision and recall as introduced by \cite{precision_recall}.

\subsection{Full pipeline}
\label{sec:full_pipeline}

We test in this subsection the capacity of our system to generate quality images. We first showcase some of the best $256 \times 256$ images we obtain from engineered prompts in \autoref{fig:best_images}. Our method is able to generate various scenes with accurate text-image alignments. The sky and the ground textures are remarkably well-depicted and the model approximately identifies when and where a natural shadow is necessary. Nevertheless, we observe that our method seems to struggle to generate high-level features such as paws, legs, and faces of animals.

\begin{figure}[h]
    \centering
    \includegraphics[width=1\linewidth]{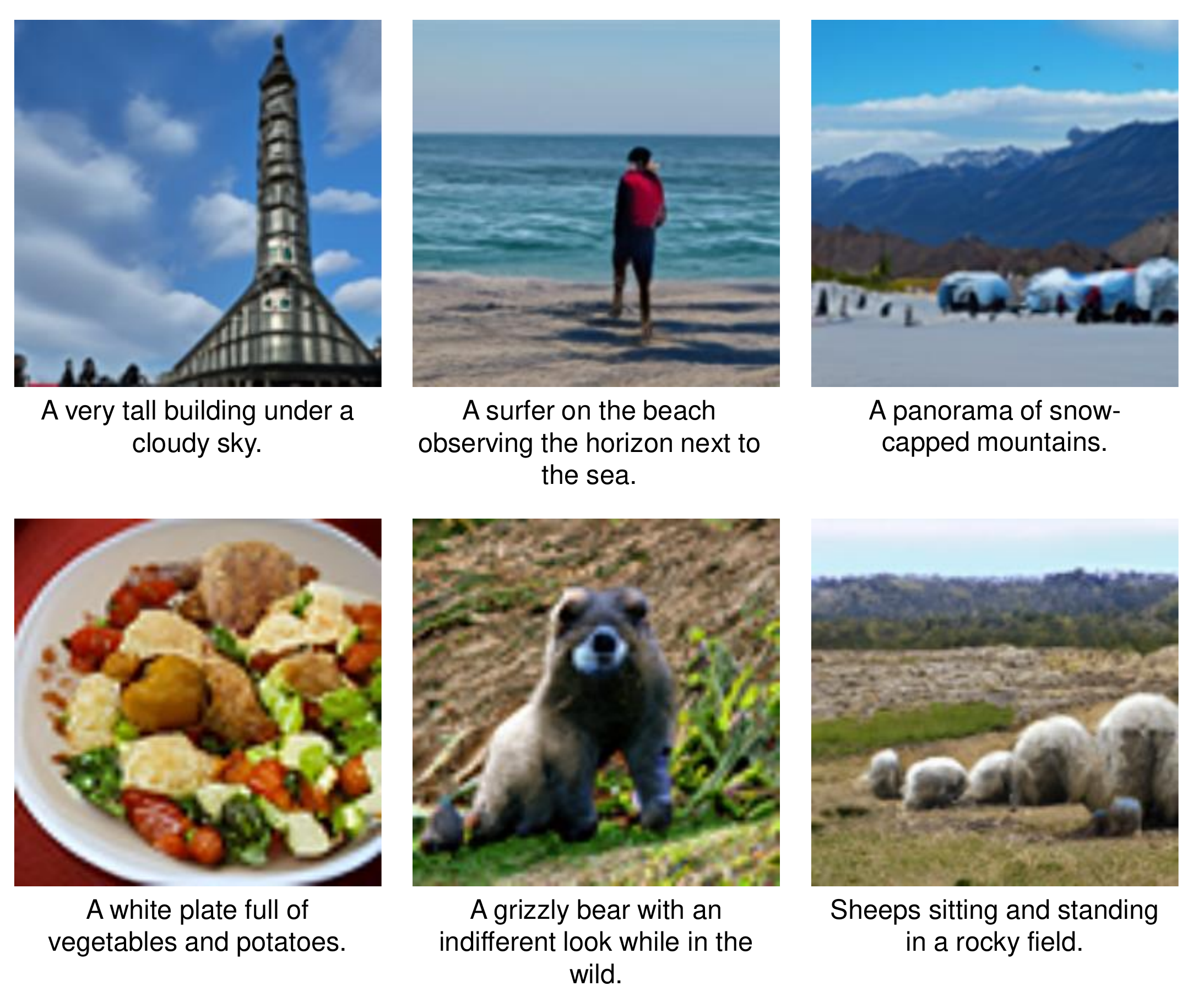}
    \caption{Sample of the $256 \times 256$ images generated by our method.}
    \label{fig:best_images}
\end{figure}

\begin{figure}[h]
\centering
\begin{minipage}[t]{.49\textwidth}
  \centering
  \includegraphics[width=1\linewidth]{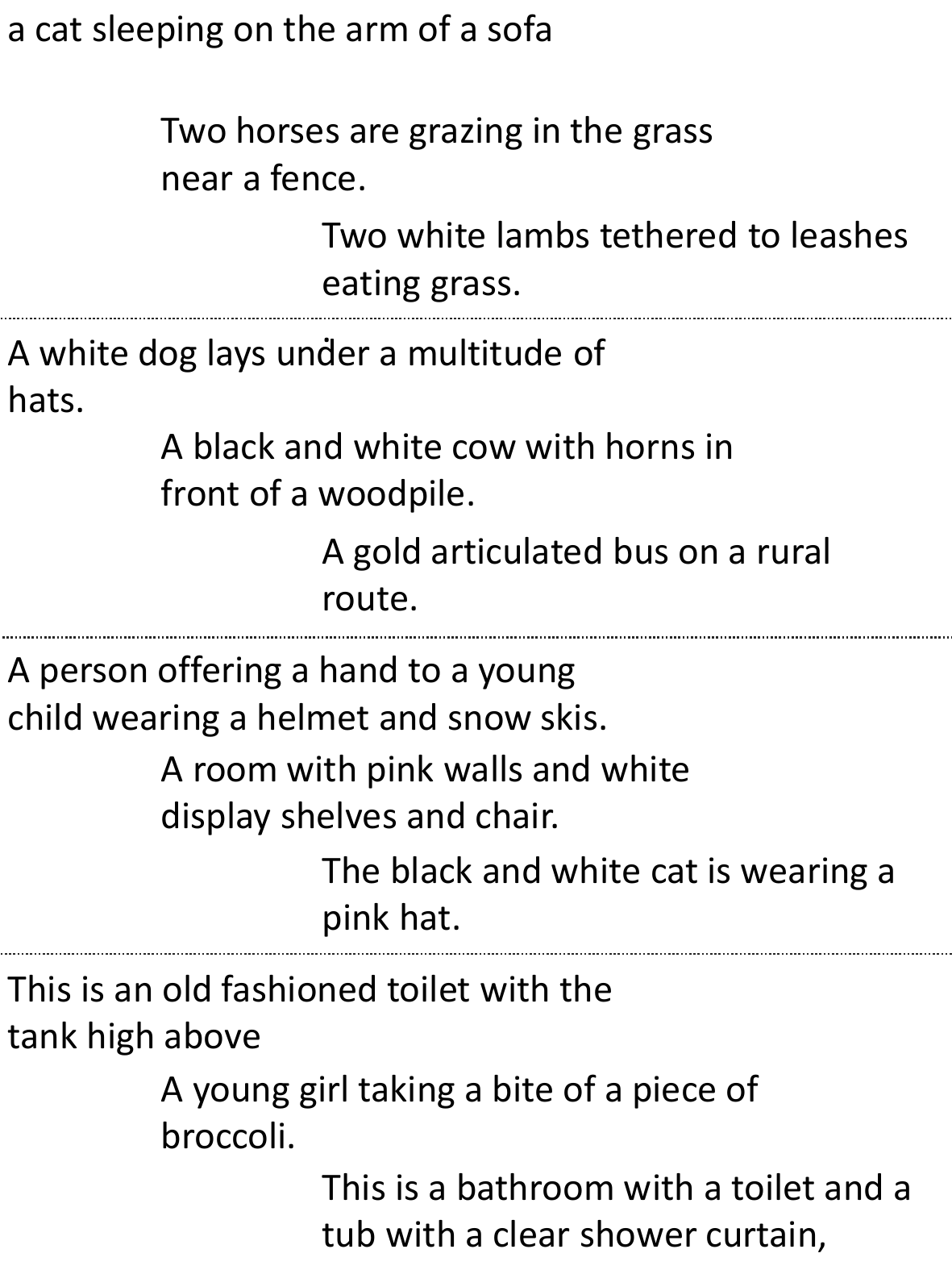}
\end{minipage}
\hspace{0.2em}
\begin{minipage}[t]{.49\textwidth}
  \centering
  \includegraphics[width=1\linewidth]{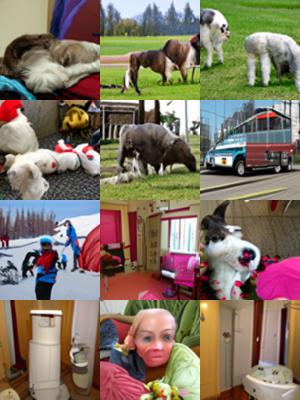}
\end{minipage}
\caption{Captions randomly drawn from the MS-COCO validation set, accompanied by the $256 \times 256$ images generated from these captions by our method. No cherry-picking at all.}
\label{fig:random_pipeline_images} 
\end{figure}

The captions and images of \autoref{fig:best_images} were carefully selected, so we now demonstrate the capacity to generate directly a single good image for any sort of caption. To do this, we consider $12$ captions randomly picked from the MS-COCO validation set \citep{coco}. Recall that all the samples of the MS-COCO dataset are held-out during training, which forces our method to learn this dataset in a zero-shot fashion. The captions and the synthetic images obtained from them are displayed in \autoref{fig:random_pipeline_images}.

We observe that the content of the captions is always present in the images, but with varying degrees of accuracy. Our system seems to be able to deal and represent small numbers since it managed to generate the right amount of horses and lambs in the top images. The colors are also respected for the different animals (lamb, cow, dog, and cat) and for the pink objects, but not for the gold bus. It appears that our model has difficulty handling complex and unusual prompts, e.g, "A white dog lays under a multitude of hats." and "The black and white cat is wearing a pink hat". Besides, as for other generative models, we notice that generating human faces seems to be difficult for our method. Nevertheless, the generated textures such as grass or fur are often of high quality. It could be explained by the fact that our super-resolution model has been trained on ImageNet, which contains many iconic images of animals in simple landscapes. However, when the upsampler has to deal with overlapping objects, the outcome can be a little blurry.

\begin{figure}[h]
    \centering
    \includegraphics[width=1\linewidth]{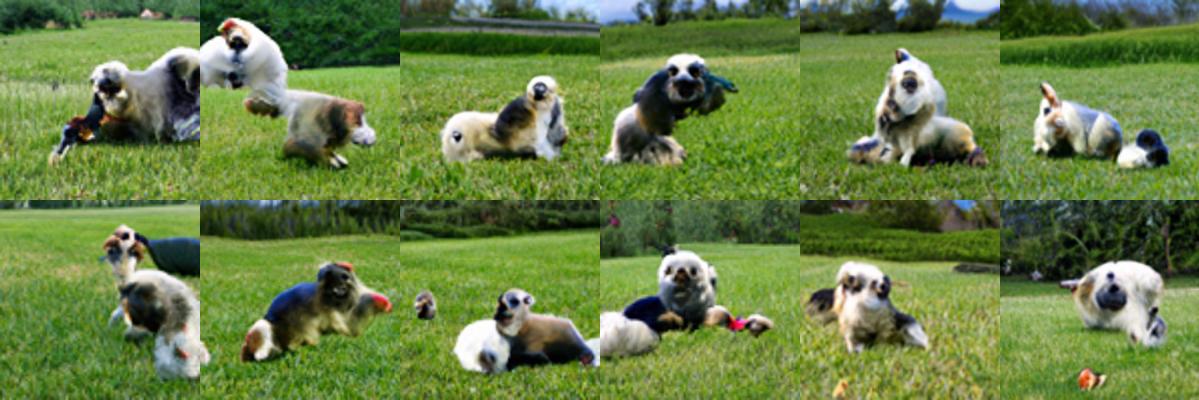}
    \caption{Random samples from our method for caption “A dog sitting on top of a grass covered field.”}
    \label{fig:dog}
\end{figure}

In addition, we demonstrate in \autoref{fig:dog} the diversity of the generated images, displaying $12$ random samples from the caption "A dog sitting on top of a grass covered field." Even for a simple caption like this one, the model produces a variety of images by trying to synthesize the dog in distinct postures with different backgrounds. We see the limits of this diversity though, since the dog is often not well depicted. More examples can be found in \autoref{sec:extra_samples}.

Finally, we compute the FID, IS, precision, and recall of 1,000 image-caption pairs randomly drawn from the MS-COCO validation set. Usually, these metrics are computed on 30,000 images, but generating images with diffusion models is long and costly, and thus we consider less images to save computational resources. We empirically observed that the FID can be drastically reduced by using more images, making this metric unsuitable for comparison with other models. However, we compute it in the same way for several ablated variations of our method, enabling us to study the contributions of the different components. \autoref{tab:metrics} compiles all the metrics obtained for each method variant. We provide some analyses of this table in \autoref{sec:clip_translator}.

\subsection{Image decoder}
\label{sec:test_image_decoder}

\begin{wrapfigure}{r}{0.45\textwidth}
    \centering
    \includegraphics[width=1\linewidth]{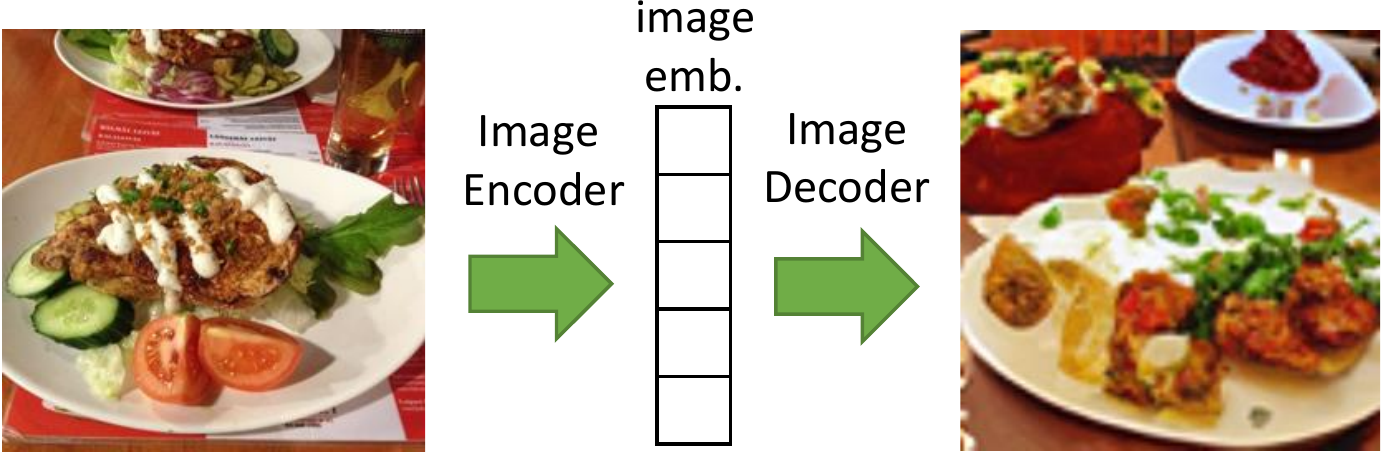}
    \caption{Image reconstruction with the image decoder.}
    \label{fig:image_reconstruction}
\end{wrapfigure}
\begin{figure}[h]
\centering
\begin{minipage}[t]{.475\textwidth}
  \centering
  \includegraphics[width=1\linewidth]{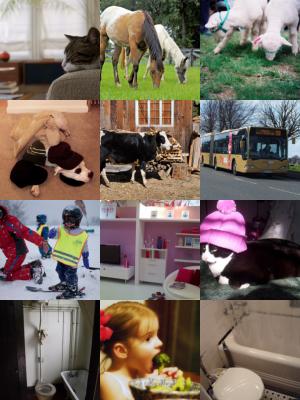}
\end{minipage}
\hspace{0.5em}
\begin{minipage}[t]{.475\textwidth}
  \centering
  \includegraphics[width=1\linewidth]{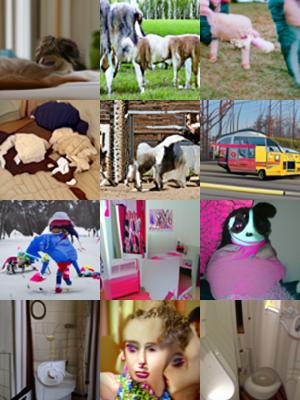}
\end{minipage}
\caption{Reconstructed images (\textbf{right}) by the image decoder conditioned on the CLIP embeddings of the original images (\textbf{left}).}
\label{fig:reconstruction} 
\end{figure}

 
We focus now on the image decoder, to see how well it can generate an image given a CLIP image embedding. To do this, we consider the same set of $12$ images used in \autoref{sec:full_pipeline} and compute their corresponding image embeddings with the CLIP image encoder. We can then reconstruct all the images using the image decoder conditioned on the embeddings. This process is illustrated by \autoref{fig:image_reconstruction}. We show the $12$ not cherry-picked reconstructed images of this set in \autoref{fig:reconstruction}, as well as the original images. Since we are using the same set of images, we can compare images generated by the image decoder only in \autoref{fig:reconstruction} to the ones obtained by the full pipeline in \autoref{fig:random_pipeline_images}. We observe that the image compositions are more similar to the original images when we directly use the CLIP image embeddings, with objects being often in the same position as in the original images. This higher fidelity could demonstrate that CLIP image embeddings capture generally more details than captions, such as the location and posture of the elements of the image. This hypothesis is also put forward by \cite{clip_see}. Nonetheless, it also frequently results in less coherent images, where it seems that the image decoder is trying to incorporate too many little details in the image. Finally and in the same way as for the full pipeline, we compute our different metrics for 1,000 generated images conditioned on the CLIP image embeddings corresponding to the same image-caption pairs used in \autoref{sec:full_pipeline}.

\subsection{CLIP translator}
\label{sec:clip_translator}

In the two previous subsections, we analyzed the differences between outputs of the full pipeline and the image decoder. The CLIP translator plays an important role since it is supposed to reduce these differences. Obviously, a caption and, consequently, its related text embedding will be always less rich than an image embedding computed directly on the image. But still the CLIP translator has to ensure that all the details provided by the caption are well translated into visual features on the images. Therefore, a modest way to understand its impact is simply to remove it from the pipeline. This means giving directly the CLIP text embeddings to the image decoder instead of first translating them to image embeddings with the CLIP translator. We then follow the same methodology as before to compute the different metrics.

\begin{table}
\centering
\begin{tabular}{lllll}
\hline
\textbf{Method variant}     & \textbf{FID} $\downarrow$ & \textbf{IS} $\uparrow$ & \textbf{Precision} $\uparrow$ & \textbf{Recall} $\uparrow$ \\ \hline \hline
Full pipeline      &  65.9   &  11.6  &    0.804       &   0.307     \\
Image decoder only &  62.8   &  11.8  &    0.696       &    0.676    \\
No CLIP translator &  124   &  7.75  &    0.315       &    0.340    \\ \hline
\end{tabular}
\caption{Summary of the different metrics. They are computed on 1,000 synthetic images only, making them difficult to compare with other works. The different variations of our method are described in \autoref{sec:testing}}
\label{tab:metrics}
\end{table}

Now that we have estimated the values of the different metrics for each method variant (see \autoref{tab:metrics}), we can try to interpret them. The lowest FID is obtained when only the image decoder is used, which is consistent with the fact that a caption captures less information about the image than the image embedding computed by the CLIP image encoder. Nevertheless, this value is very close to the one obtained with the full pipeline, which indicates that the CLIP text encoder and CLIP translator are doing a good job to transfer the textual content of the caption to the image decoder. This is confirmed by the FID obtained when the CLIP translator is removed from the pipeline, which is almost twice as large as the others. Therefore, even if the CLIP text and image embeddings are already closed for similar contents, a CLIP translator is necessary to switch from one to the other.

As explained above, the FID is the most common metric to measure the performances of image generative models, so we are not going to do much analysis on the other metrics. We note that the IS behaves similarly to the FID here. However, the full pipeline variant obtained the highest precision but the lowest recall. One interpretation which is linked to what we noticed in \autoref{sec:test_image_decoder} is that using a caption to describe an image simplifies its content, involving that the generated images will more likely fall in the support of the real images distribution. It hence induces a higher precision. But on the other hand, the generated images will only cover a small portion of the support of the real images distribution, causing this time a lower recall. More details about precision and recall can be found in \cite{precision_recall}.

\subsection{Upsampler}

Finally, we devote this subsection to the assessment of the upsampler, which increases the image resolution from $64 \times 64$ to $256 \times 256$. As mentioned earlier, we use a super-resolution diffusion model already trained on ImageNet. In contrast, the other state-of-the-art DALL-E 2 and Imagen train their own upsamplers, which are additionally conditioned on the embedding. We have however no way to compare our $64 \times 64$ generated images to the ones produced by DALL-E 2 and Imagen, since they provide only the final $1024 \times 1024$ images. We can still assume that it is tricky to model complex scenes and objects in the low resolution regime, involving the necessity to use powerful upsamplers handling high-level features.

\begin{figure}
    \centering
    \includegraphics[width=1\linewidth]{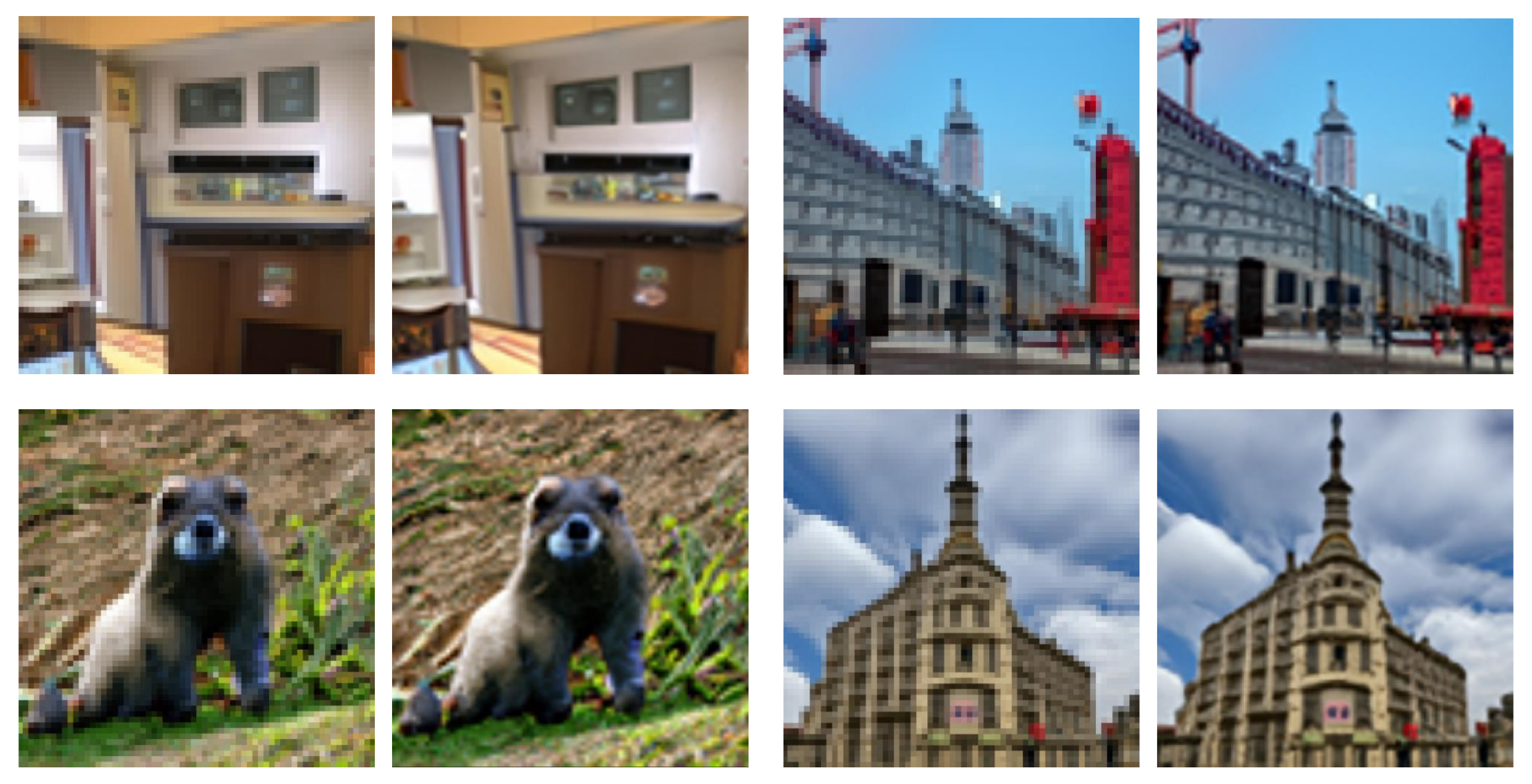}
    \caption{Differences between $64 \times 64$ and $256 \times 256$ images, the latter being obtained using our upsampler model.}
    \label{fig:upsampler}
\end{figure}

We show in \autoref{fig:upsampler} pairs of images before and after being inputted to our upsampler. We can observe that the $256 \times 256$ images are not really sharp, often with some blurry parts. But as mentioned above, some textures are nevertheless well enhanced by the upsampler, such as the cloudy sky in the right bottom image. This upsampler is then still a useful one as its implementation is available.

\section{Exploring the properties of CLIP embeddings }
\label{sec:clip_exploring}


We have shown in \autoref{sec:clip} that CLIP embeddings have many useful properties. However, it has not yet been demonstrated if these embeddings can be manipulated and combined in their continuous space representation. In particular, we are referring to the seminal work of \cite{mikolov2013linguistic} which exhibited that linguistic regularities appear in the vector-space of word embeddings, enabling them to perform some vector operations. For instance, they showed that the vector corresponding to "Man + Queen - Woman" resulted in a vector very close to "King". We would like therefore to see in this section if CLIP embeddings also learn relationships between concepts and if it allows us to perform simple vector-oriented reasoning.

\subsection{Image embeddings}

\begin{figure}
    \centering
    \includegraphics[width=0.9\linewidth]{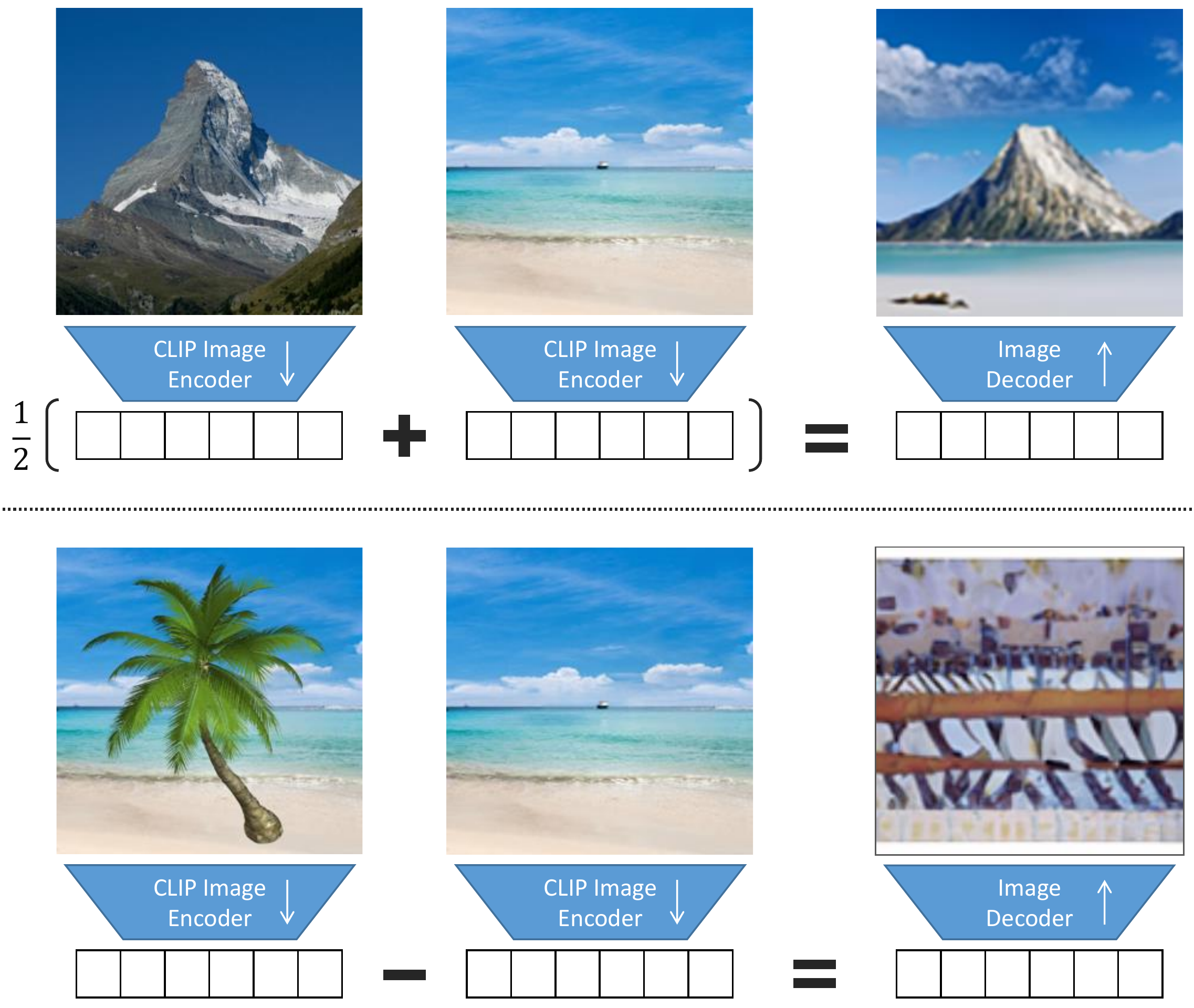}
    \caption{Simple vector operations on the CLIP image embeddings. The average of two embeddings yields a combination of the content of the two images (\textbf{top}). However, subtracting one part of the image leads to embeddings which are out of the training distribution, resulting in peculiar images (\textbf{bottom}).}
    \label{fig:img_algebra}
\end{figure}

We begin by investigating the properties of the CLIP image embeddings. To keep our analyses independent from the CLIP translator, we use only the image decoder and not the full pipeline. Our first experiment consists in testing the semantic robustness of the image embeddings under simple vector operations such as additions and subtractions. To do this, we compute the image embeddings of two different images with the help of the CLIP image encoder. Next, we perform 1) a vector average and 2) a vector difference between these two embeddings. We then use the image decoder to generate the images corresponding to the embeddings calculated. The results are displayed in \autoref{fig:img_algebra}. We observe that taking the average works well and leads to a satisfactory outcome combining the mountain of the first image and the beach of the second one. The subtraction, on the other hand, results in an odd image, when it was supposed to simply represent a palm tree. It could be explained by the fact that the embedding obtained after the subtraction could end up in an unseen area of the embedding space.

\begin{figure}[H]
    \centering
    \includegraphics[width=1\linewidth]{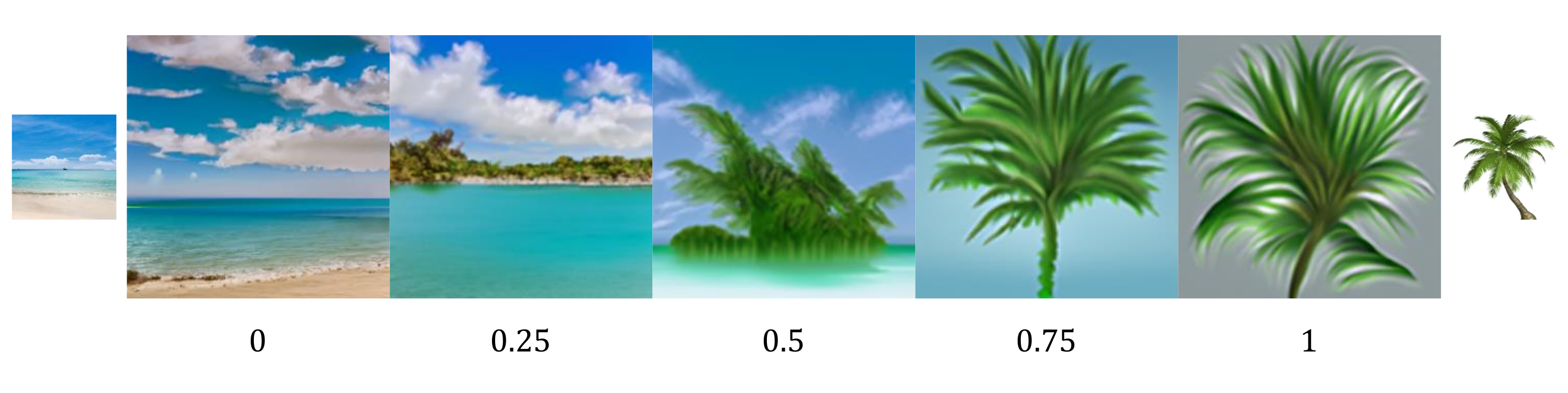}
    \caption{Linear interpolation between two image embeddings. The interpolation coefficient is indicated on the bottom. The rightmost and leftmost images are the original ones.}
    \label{fig:interpolation}
\end{figure}

We push the analysis further by considering different points of the linear interpolation between two image embeddings in \autoref{fig:interpolation}.
We observe intermediate variations of the images content. At interpolation coefficient equals to $0.25$, the beach gets some greenery, whereas at $0.75$ only the blue background remains from the image, replaced by the palm tree. The transition between one image to another is therefore rather smooth.

\subsection{Text embeddings}

\begin{figure}[h]
    \centering
    \includegraphics[width=1\linewidth]{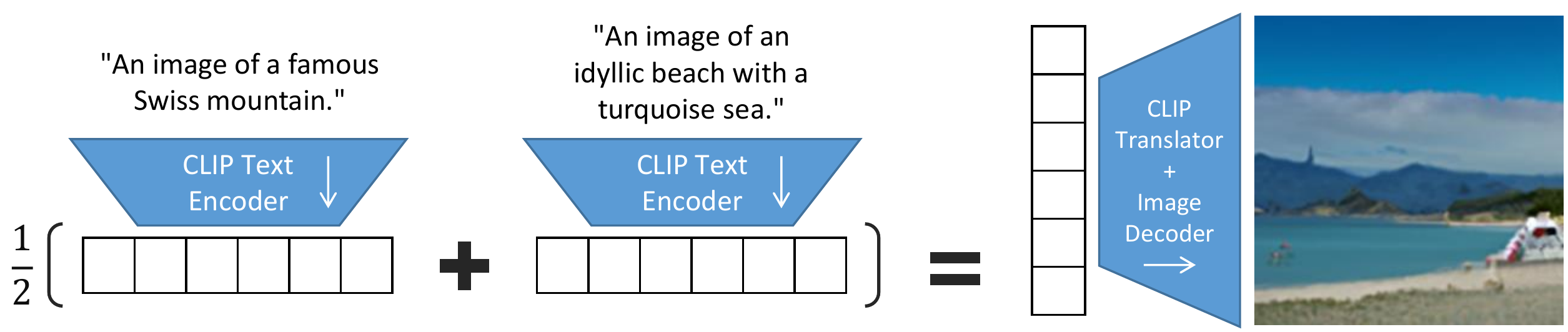}
    \caption{Combination of the content of two captions by an embedding average.}
    \label{fig:txt_algebra}
\end{figure}

We now focus on the text embeddings to combine contents of different captions. We start by two captions which are encoded and then averaged. We use the CLIP translator to translate the resulting text embedding to an image embedding and decode it with our image decoder. We obtain the image of \autoref{fig:txt_algebra}. Similarly to image embeddings, we obtain an image including the content of both captions, as if the two sentences had simply been concatenated to form a single caption.

\begin{figure}[h]
    \centering
    \includegraphics[width=1\linewidth]{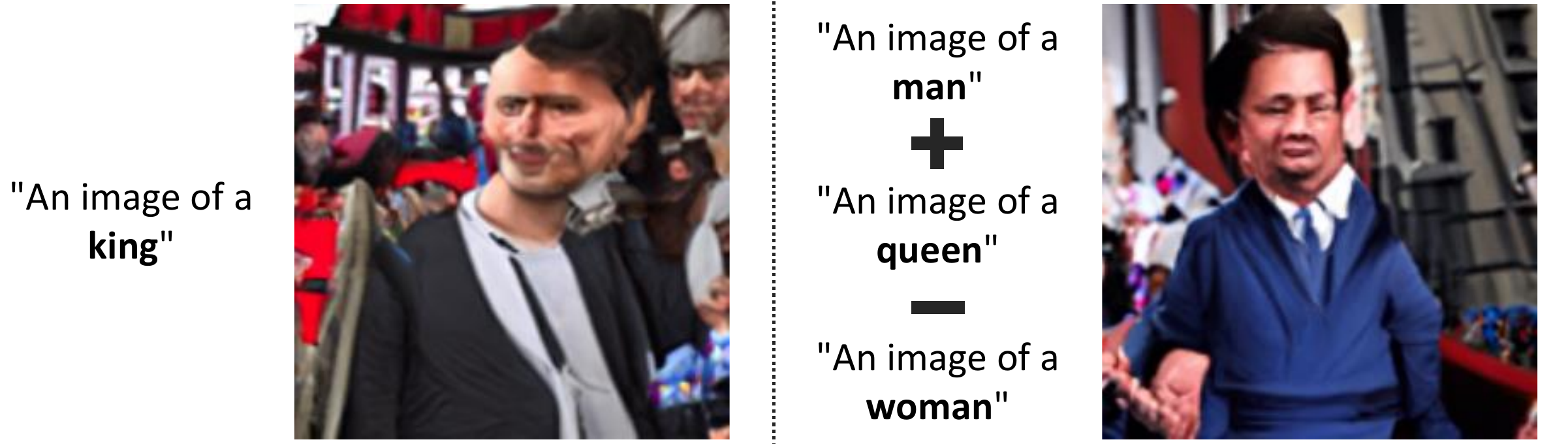}
    \caption{The image decoded from the caption "An image of a king" (\textbf{left}). The image decoded from the displayed vector operation on text embeddings (\textbf{right}). The two images depict modern kings.}
    \label{fig:king}
\end{figure}

Our final experiment on CLIP embeddings consists of reproducing the famous example given by \cite{mikolov2013linguistic}. We expect that the embedding resulting from the vector operation "An image of a man" + "An image of a queen" - "An image of a woman" is decoded in an image of a king. The image obtained, as well as an image corresponding to the caption "An image of a king", can be seen in \autoref{fig:king}. We observe that the two images are similar, portraying both a man in a suit\footnote{In particular, we generated $16$ images per embedding, and we could not distinguish which samples came from which embedding (see \autoref{sec:more_image_king}).}.
Despite the fact that our first perception of a king generally consists of a man with a crown sitting on a throne, modern kings are now more often men in suits, such as the King of Spain. Our dataset must therefore contain more kings of this type.

\section{Image guidance}
\label{sec:results_image_guidance}

\begin{figure}[h]
    \centering
    \includegraphics[width=1\linewidth]{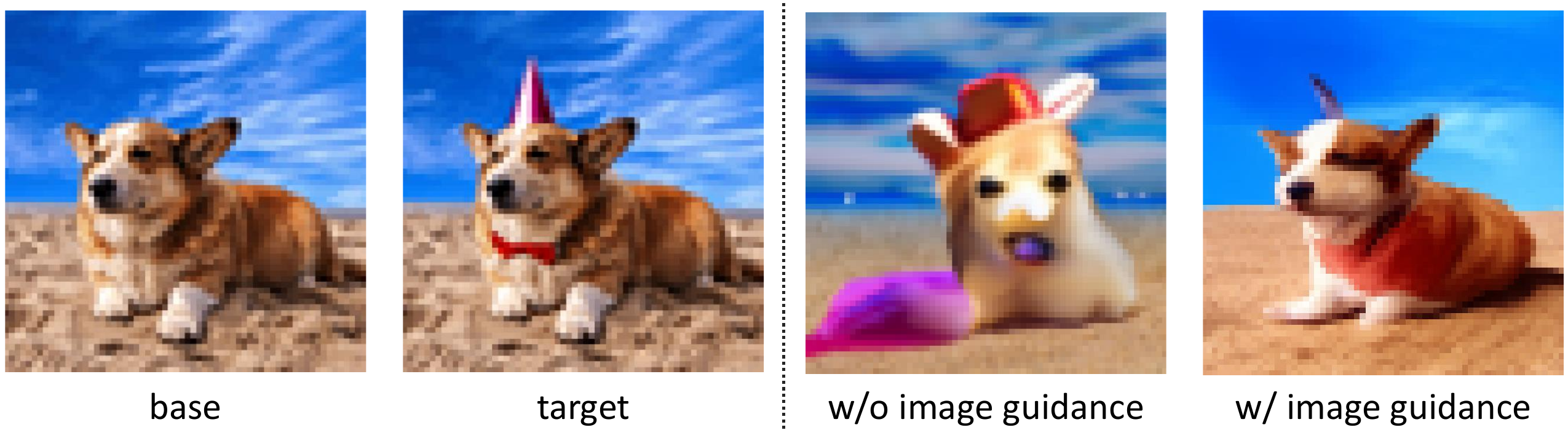}
    \caption{Illustration of the utility of image guidance, used here with the base image. Reproducing the content of the target image from its image embedding is easier with image guidance than without. All images are $64 \times 64$.}
    \label{fig:corgi}
\end{figure}

In \autoref{sec:image_guidance}, we introduced a new guidance method named image guidance. It consists in guiding the image generation process towards the direction of another image of our choice. In this section, we conduct a preliminary experiment to see what this method is capable of doing. In particular, we would like to know if it can help the generating process to produce higher quality images. To test this, we consider an image of a corgi lying on the beach, as well as a variation of this image where the corgi is additionally wearing a purple party hat and a red bow tie, as depicted in \autoref{fig:corgi}. The first image is used for image guidance, and the second one is encoded into a CLIP image embedding. We then use our image decoder conditioned on the image embedding with the aim of generating images similar to the second image, i.e., an image of a corgi with a hat and a bow tie. We test this with and without image guidance. The image guidance scale is set to $0.005$, and is linearly decayed over the timesteps. The outputs are displayed in \autoref{fig:corgi}.

We observe that the image generated with image guidance is way more consistent than the one without. It seems that image guidance can help the diffusion model by providing it a base image on which it can build on, allowing the diffusion model to better reproduce small details, such as the position and color of the hat and bow tie. Nevertheless, we consider here only the image decoder. In real-life application contexts, we do not have access to the target image, nor its embedding. 

\begin{figure}[h]
    \centering
    \includegraphics[width=1\linewidth]{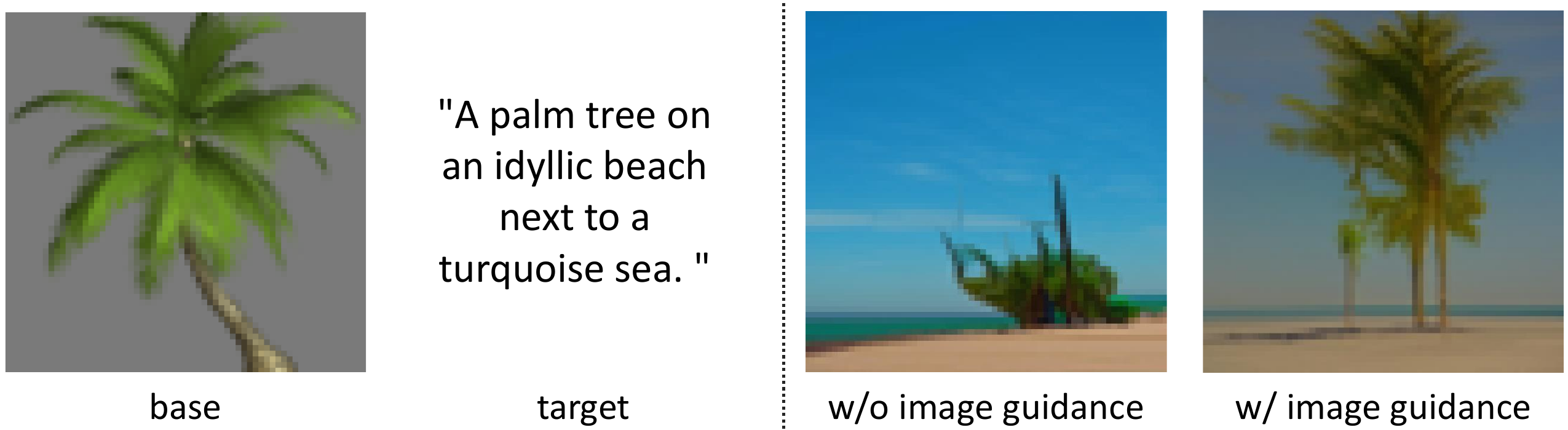}
    \caption{The base image with image guidance helps to obtain better text-image alignments from a given target caption.}
    \label{fig:image_guidance_tree}
\end{figure}

Instead, we can use our full pipeline to obtain the embedding of a textual description of the target image. Then we can decode the embedding with and without the help of image guidance as previously. The elements used in this process are shown in \autoref{fig:image_guidance_tree}, as well as the generated images (more of them can be found in \autoref{sec:more_image_guidance}). We notice here again that the base image assists the diffusion model to correctly interpret the embedding. Indeed, the image and especially the palm tree produced with image guidance is more the outcome we expected for the given caption than the image obtained without image guidance. It then appears that image guidance is a useful tool to help the generating process, by providing extra information, taking the form of an image, about the desired outcome.

However, this method is obviously greatly limited. Firstly, an appropriate $64 \times 64$ image must be found to accordingly serve as a base image, along with the proper image guidance scale. Besides, image guidance induces an important mismatch between training and sampling as the base image is added at each timestep, but this is the case for other guidance methods as well. For these reasons above, we struggled to produce quality images to illustrate the effectiveness of image guidance, in particular for image inpainting. It might be then interesting to see how this method performs with better models such as DALL-E 2 and Imagen.

\chapter{Discussion}
\label{sec:discussion}

Now that we have conducted several experiments, we discuss the results obtained, and especially the possibilities and limitations of our method. We start by comparing and explaining the differences between our approach and the related works. Next, we propose some future directions that could be taken to continue and improve this work. Finally, because generative models can have a deep impact on society, we briefly mention some ethical implications of these models.

\section{Comparison with related works}
\label{sec:comparison}

The quality of our synthetic images is difficult to compare directly with the ones of DALL-E 2 and Imagen. These two models consider way more resources and therefore obtain highly realistic images. In particular, we list here the differences between our method and DALL-E 2 which could have an impact on the final outcome:
\begin{itemize}
    \item \textbf{Dataset size}: DALL-E 2 uses 650M image-caption pairs, whereas we only use 12M.
    \item \textbf{Diffusion model size}: The diffusion model employed by DALL-E 2 to decode CLIP image embeddings has a larger architecture than ours, e.g., $512$ base channels, only $256$ for us. It results in 3.5B and 0.5B learnable parameters respectively.
    \item \textbf{CLIP encoders size}: We use smaller pre-trained CLIP encoders compared to DALL-E 2. For the CLIP image encoder we consider the ViT-B/32. They use the ViT-L/14, and their CLIP text encoder is twice as wide and deep. The authors of Imagen showed that the size of the language model could make a considerable difference on the performance of the whole system.
    \item \textbf{Upsamplers}: We use a super-resolution model trained on ImageNet only, while the authors of DALL-E 2 train their own upsamplers on their large dataset. Nevertheless, it allows us to avoid the large computational cost of the training.
    \item \textbf{CLIP translator}: We use a small MLP model (30M parameters) for the CLIP translator, whereas DALL-E 2 uses a large diffusion model (1B parameters).
    \item \textbf{Training}: There are a few differences between the training of DALL-E 2 and ours. Firstly, they train their image decoder with a batch size of 2,048 for $2.5$ epochs. For reference, we only train our image decoder for $1$ epoch on a $54$ times smaller dataset. Moreover, we had to choose a small batch size to make the GPU able to support the heavy computations. It resulted in exploding gradients, solved by gradient clipping. It could have an impact on the performance though. Finally, they use optimization tricks and tune their hyperparameters well, which could not be replicated without further increasing computational demand.
\end{itemize}

All these differences therefore lead to significantly different performances, but also to different computational costs. Our method may not have the sample quality of DALLE-2, but it still manages to understand the content of the captions and to reproduce them with varying degrees of success, without the need for notably high computational resources. Especially, it has the same properties of DALL-E 2 which allow us to manipulate CLIP embeddings, as we have seen in \autoref{sec:clip_exploring}. It enables combining images and captions, and to obtain a deeper comprehension of the content of both text and image embeddings.
This property is not present with Imagen, which does not represent images explicitly by embeddings.
Nevertheless, the model structure of Imagen is simpler and has the benefit of leveraging a huge pre-trained language model. Its text modeling capacity is therefore higher, and the output for a given caption is a sequence of vectors instead of a single vector as DALLE-E 2 and us, capturing potentially more content subtleties.

Another advantage of our method and DALL-E 2 over Imagen, is the possibility to use distinct datasets to train the image decoder and the CLIP translator. Indeed, the image decoder can use any image dataset during training, not only captioned images. Note that this is the reason why we have included ImageNet in our dataset pool. The CLIP translator, on the other hand, needs pairs of text and image embeddings, which are lighter to store than images and captions. Thus, it enables to consider more easily massive datasets such as LAION \citep{LAION}, whose authors provide the CLIP embeddings for each image and caption.

\section{Future work}

The design and training of text-to-image generative models are based on many heuristics. Thus, an infinite number of experiments can be done by just trying various combinations of the different hyperparameters. However, recent observations exhibit that just scaling the size of the dataset and the model leads almost all the time to performance improvements. It results then in the traditional trade-off between computational cost and model performance.
Accordingly, a longer and more careful training of our diffusion model should definitely improve its effectiveness. We could also consider the larger CLIP encoders which are publicly available and pre-trained. In particular, Imagen demonstrated that huge language models were the key to enhance sample quality and text-image alignment. Furthermore, training our own upsamplers could also enable us to produce better high-level image features.

We conducted several experiments in this work. We would have liked to test out more ideas, but the sampling time of our method, due to the sequential nature of the diffusion models, restricted us. We propose therefore as a future work to implement and test DDIM \citep{DDIM}, in order to reduce the number of steps required during sampling. It would also allow to compute the FID more efficiently and to experiment in depth the capacities of image guidance.

Finally, as mentioned above, the DALL-E 2 pipeline is complex and requires the juxtaposition of different models. Leveraging instead a huge pre-trained language model as Imagen does reduces this complexity. We hence suggest trying to incorporate this element to our codebase. It would be very easy to implement as the only modification to our diffusion model is to give as input the embedding obtained from the language model instead of the one from the CLIP translator. If the upsamplers are also trained, it would result in a replication of Imagen, which could probably synthesize greater images than our current implementation.

\section{Societal impacts}

Outputs of large-scale text-to-image models have an impact on society, with direct repercussions on individuals. In this section, we propose a brief review of the principal issues and opportunities offered by these models.

\textbf{Energy consumption.} Deep neural networks and especially deep generative models often require high computational resources, consisting mainly of hours or even days of training on modern tensor processing hardware. It consumes a considerable amount of energy whose production releases CO$_2$ emission in the atmosphere\footnote{The energy produced by renewables does not emit CO$_2$, however no country in the world attains 100\% renewable energy for now.}, contributing to climate change. \cite{EnergyDL} estimated that the cost in terms of kgCO$_2$ emission of performing neural architecture search could attain 313,078 CO$_2$e, the equivalent of 8.6 years of average energy consumption for an American. \cite{reportEnergy} therefore encourages the machine learning community to systematically report the energy consumption of their models. Consequently, we use \textit{cumulator}, a tool developed at EPFL by \cite{cumulator}, to quantify the carbon footprint of the training of our models. We get $14.2$ kgCO$_2$ to train the image decoder and $0.2$ kgCO$_2$ for the CLIP translator. The amount for the image decoder is considerable and reflects the issues of these huge models, which require colossal amounts of computational resources. Nevertheless, we observe that choosing a simple MLP model for the CLIP translator was judicious since it avoids the heavy computations required by the prior diffusion model of DALL-E 2.

\textbf{Dataset bias.} It has been well-established that massive web-scraped datasets contain inherent bias \citep{pyrrhic, revise}, mirroring harmful racism and gender stereotypes among other things. In addition, \cite{stereotypes_multimodal} showed that one of these image-text datasets, LAION \citep{LAION}, includes explicit images of rape and pornography. Therefore, text-to-image generation models trained on these datasets can reproduce problematic and prejudicial contents \citep{glide}, which results in the amplification of these issues. Even if our method is not really concerned since the datasets we use are mostly filtered and curated, it is important to raise awareness to prevent detrimental impacts on people who are already subject to discrimination. We therefore strongly encourage future works on text-to-image generation to take into account these considerations, by implementing for example suggestions from \cite{model_cards} and \cite{gebru_datasheets}.

\textbf{Malicious uses.} There is a high potential for misuse when it comes to generative models, as they can be used to generate deepfakes or violent images for harmful downstream applications \citep{deep_fakes}. The photorealistic but synthetic images produced by DALL-E 2 or Imagen could lead to the public being misinformed, or even manipulated. It hence could considerably reduce the trust that individuals have when they see images on the internet. For now, these large text-to-image generation models are not available to the public community as their authors are searching for ways to alleviate these problematic misuses. But even in the hypothesis that solutions are found, it may not be sufficient to prevent all types of adversarial attacks, such as the one shown by \cite{hidden_vocabulary} on DALL-E 2.

\textbf{Beneficial uses.} Of course, text-to-image generative models do not only have disadvantages for society. Their ability to create high-quality images of any kind simply based on textual descriptions opens the door to numerous, beneficial and artistic applications.
They can be a tool to enhance human creativity and which could be available to anyone. In particular, artists could leverage these models to explore and boost their imagination. Moreover, these models can contribute to social causes, e.g., by synthesizing pictures depicting the dramatic effects of climate change, with the aim to raise public awareness \citep{climate_change}.

\chapter{Conclusion}
\label{sec:conclusion}

Throughout this thesis, we investigated diffusion models for text-to-image generation. We started by reviewing the different elements which contributed to the recent progresses of text-to-image generative models, delving into diffusion models literature and the guidance methods employed to enhance them. We then implemented our own model to generate images from textual descriptions. It is an adapted replication of the state-of-the-art model DALL-E 2, which however requires considerably less computational resources to train than its counterpart. We make this implementation available to the public community.

Next, we experimented with our model to understand what are the components which make text-to-image models so effective. We discovered that certain types of images were easier to generate. We also found out that the embeddings representing texts and images exhibited semantic regularities, allowing us to perform vector operations to manipulate and combine the content of different texts and images. In addition to this, we introduced a new guidance method named image guidance. We demonstrated that image guidance has the potential to help text-to-image models to produce images of higher quality and with better text-image alignments.

Moreover, we compared our method to other state-of-the-art models in text-to-image generation. We noticed that a few improvements could be obtained by scaling the size of our different models and datasets, but at the cost of a greater computational load. We also mentioned the societal issues raised by the use of larger models and datasets. We proposed future directions for this project which can be implemented without too much effort. In particular, replicating a model such as Imagen by integrating a larger pre-trained language model could have an important impact on the performances. We finally strongly encourage to conduct further experiments, in order to contribute to the currently thin body of knowledge the AI community has on these models.


\cleardoublepage
\phantomsection


\bibliography{thesis}

\appendix

\chapter{Extra random samples}
\label{sec:extra_samples}

\begin{figure}[h]
    \centering
    \includegraphics[width=1\linewidth]{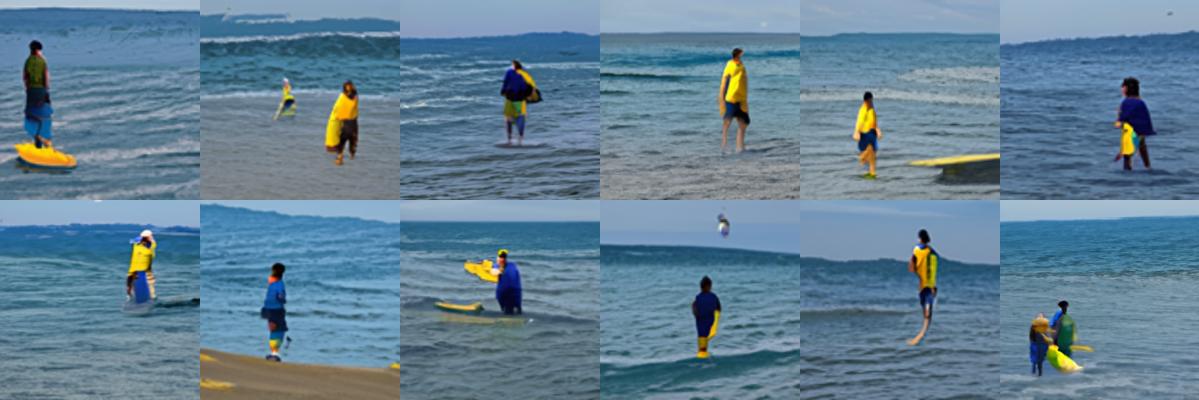}
    \caption{Random samples from our method for caption “A man holding a yellow surfboard walking into the ocean.”}
    \label{fig:surfer}
\end{figure}

\begin{figure}[h]
    \centering
    \includegraphics[width=1\linewidth]{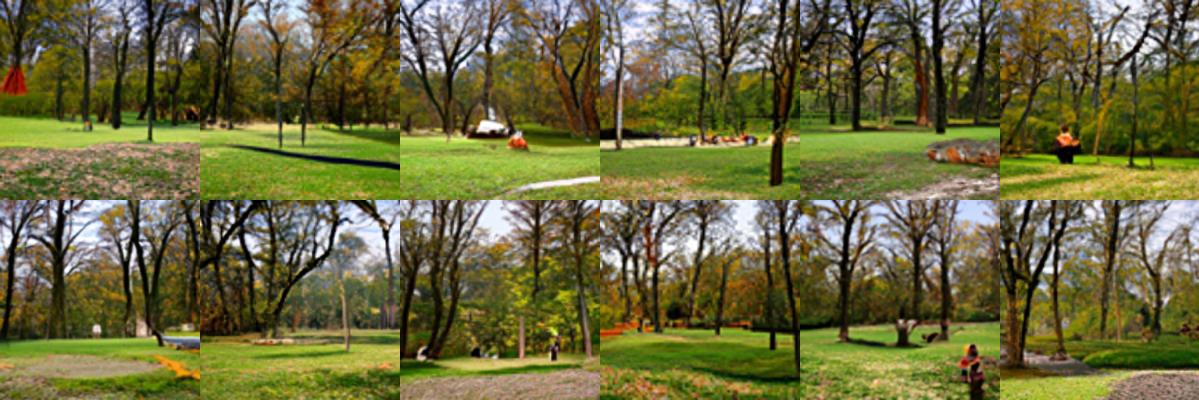}
    \caption{Random samples from our method for caption “A park covered in leaves filled with lots of trees.”}
    \label{fig:park}
\end{figure}

\begin{figure}[h]
    \centering
    \includegraphics[width=1\linewidth]{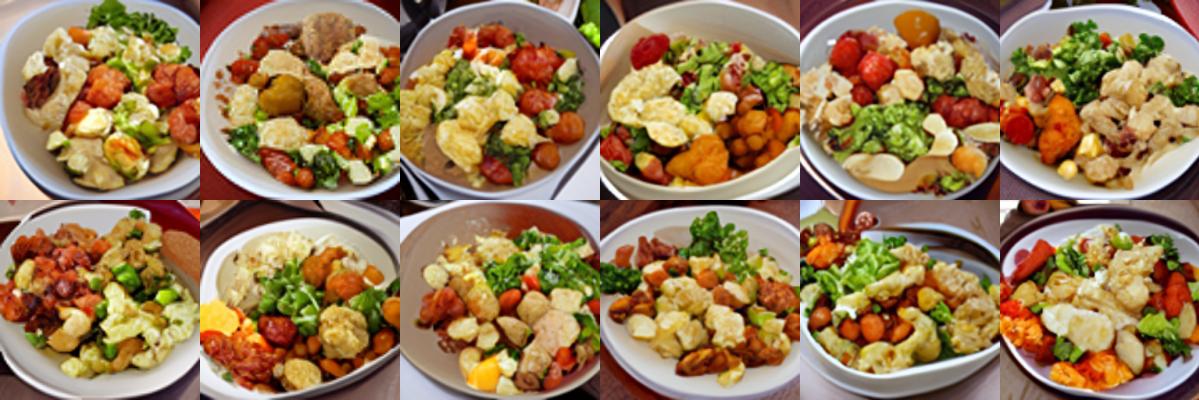}
    \caption{Random samples from our method for caption “A white plate full of vegetables and potatoes.”}
    \label{fig:plate}
\end{figure}

\begin{figure}[h]
    \centering
    \includegraphics[width=1\linewidth]{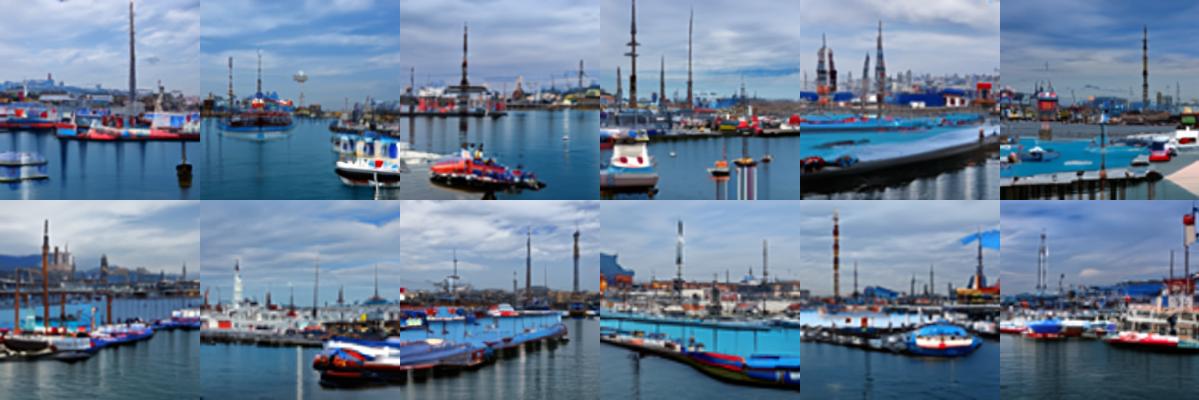}
    \caption{Random samples from our method for caption “Many boats lined up in the harbour.”}
    \label{fig:harbour}
\end{figure}

\begin{figure}[h]
    \centering
    \includegraphics[width=1\linewidth]{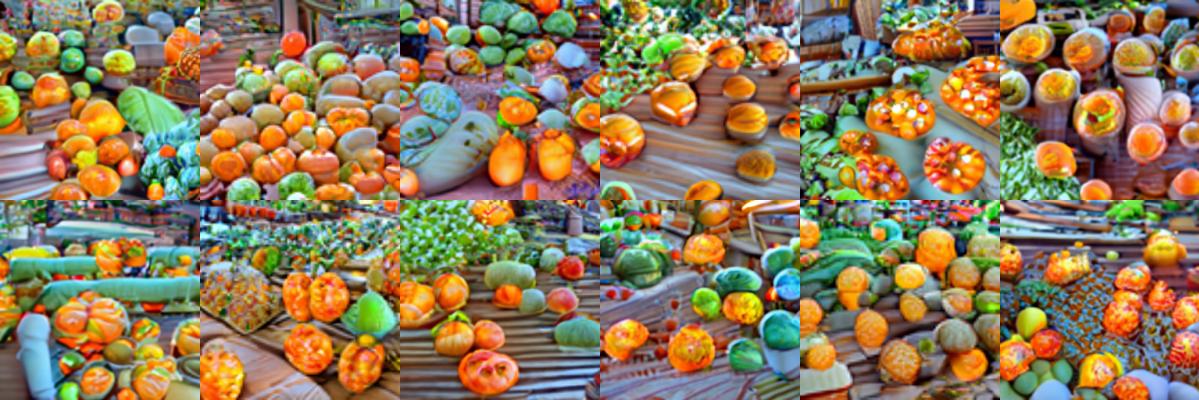}
    \caption{Random samples from our method for caption “A market fruit display of oranges.”}
    \label{fig:oranges}
\end{figure}

\chapter{More images of kings}
\label{sec:more_image_king}

\begin{figure}[h]
\centering
\begin{subfigure}[b]{1\textwidth}
   \includegraphics[width=1\linewidth]{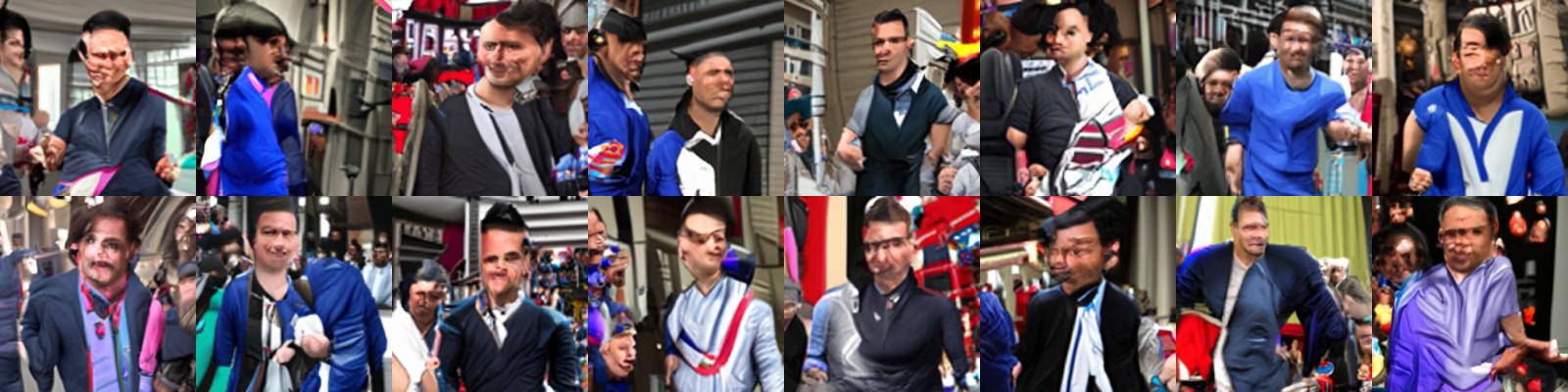}
   \caption{“An image of a king.”}
   \label{fig:king_appendix} 
\end{subfigure}

\begin{subfigure}[b]{1\textwidth}
   \includegraphics[width=1\linewidth]{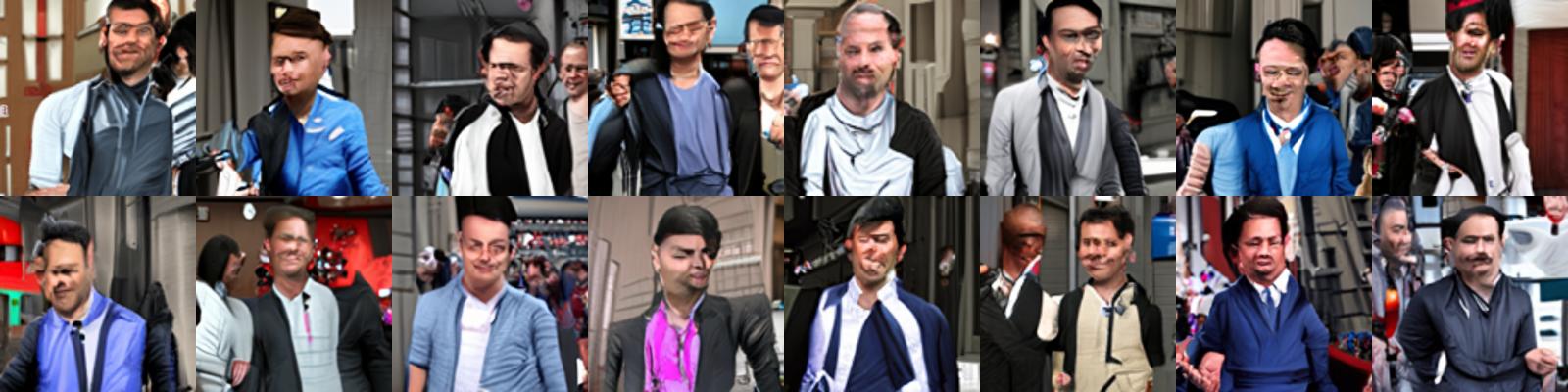}
   \caption{"An image of a man." + "An image of a queen." - "An image of a woman."}
   \label{fig:operation_king_appendix}
\end{subfigure}

\caption[]{All the generated images from the experiment described in \autoref{sec:clip_exploring}, consisting of generating images of kings with and without a CLIP text embedding operation. The two image distributions are very similar, depicting men in suits.}
\end{figure}

\chapter{More on image guidance}
\label{sec:more_image_guidance}

\begin{figure}[h]
    \centering
    \includegraphics[width=1\linewidth]{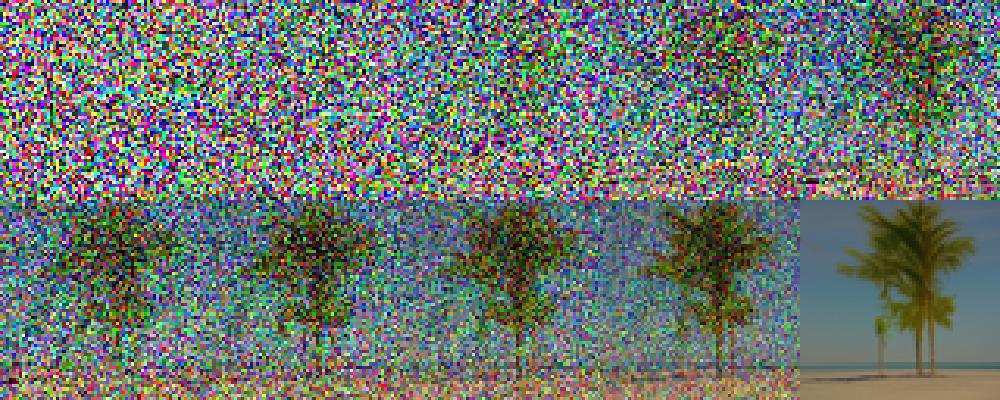}
    \caption{The latent images of different timesteps (multiples of 100) for the experiment on image guidance described in \autoref{sec:image_guidance}. We observe that the base image used for image guidance is not perceptible during the whole reverse process, but still is able to guide the generation in another direction than without it.}
    \label{fig:transition}
\end{figure}

\begin{figure}[h]
\centering
\begin{subfigure}[b]{1\textwidth}
   \includegraphics[width=1\linewidth]{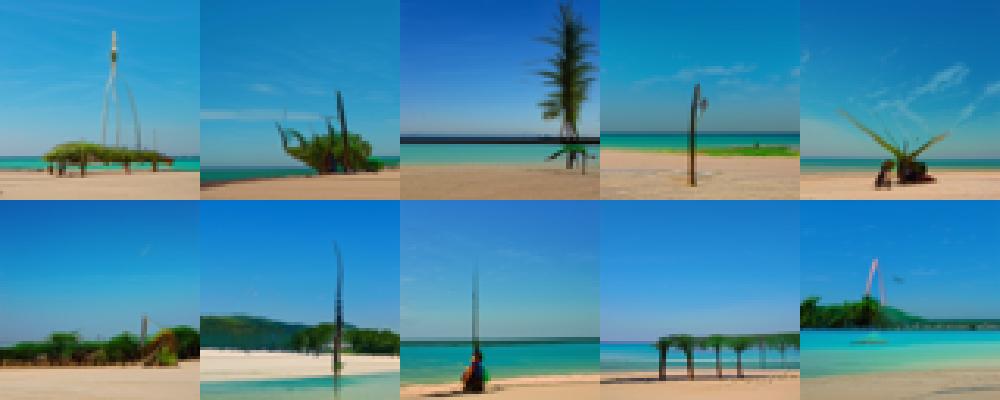}
   \caption{Without image guidance}
   \label{fig:all_trees_without} 
\end{subfigure}
\begin{subfigure}[b]{1\textwidth}
   \includegraphics[width=1\linewidth]{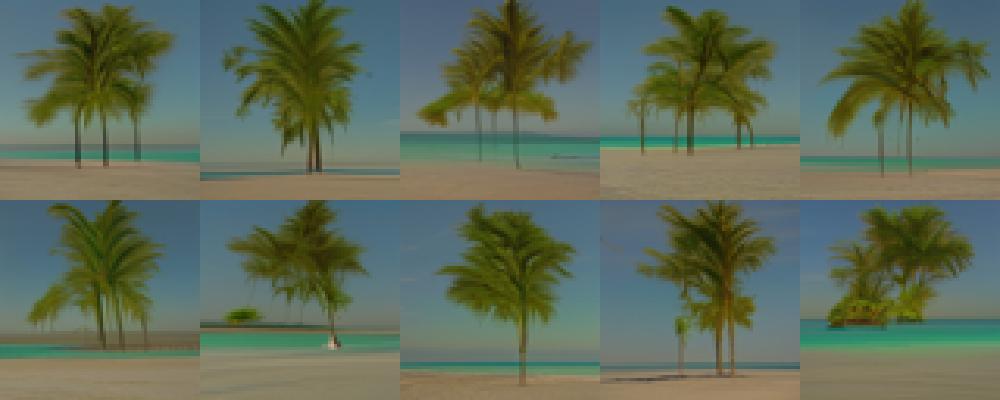}
   \caption{With image guidance}
   \label{fig:all_trees_with}
\end{subfigure}

\caption[]{All the generated images from the experiment on image guidance described in \autoref{sec:image_guidance}. We notice that the images generated with image guidance produce palm trees of better quality than without.}
\end{figure}

%
%

\end{document}